\documentclass[english]{llncs}
\usepackage[T1]{fontenc}
\usepackage{color}
\usepackage{array}
\usepackage{verbatim}
\usepackage{rotating}
\usepackage{url}
\usepackage{multirow}
\usepackage{amsmath}
\usepackage{stackrel}
\usepackage{graphicx}

\makeatletter

\providecommand{\tabularnewline}{\\}

\everymath{\displaystyle}

\makeatother

\usepackage{babel}
\begin{document}

\title{Fiber-Flux Diffusion Density for White Matter Tracts Analysis: Application to
Mild Anomalies Localization in Contact Sports Players}


\author{Itay Benou\inst{1,3} \and Alon Friedman\inst{2,3,4} \and Tammy Riklin Raviv\inst{1,3}}

\institute{Department of Electrical Engineering,\\
	Ben-Gurion University of the Negev, Beer-Sheva, Israel\\
	\and
	Department of Physiology and Cell Biology,\\
	Ben-Gurion University of the Negev, Beer-Sheva, Israel\\
	\and
	The Zlotowski Center for Neuroscience,\\
	Ben-Gurion University of the Negev, Beer-Sheva, Israel\\
	\and
	Departments of Medical Neuroscience and Brain Repair Centre,\\
	Dalhousie University, Faculty of Medicine, Halifax, Canada
	}

\maketitle
\begin{abstract}
We present the concept of fiber-flux density for locally quantifying
white matter (WM) fiber bundles. By combining scalar diffusivity measures
(e.g., fractional anisotropy) with fiber-flux measurements, we define
new local descriptors called {\it Fiber-Flux Diffusion Density}\index{Fiber-Flux Diffusion Density} 
(FFDD) vectors. Applying each descriptor throughout fiber bundles
allows along-tract coupling of a specific diffusion measure with geometrical
properties, such as fiber orientation and coherence. A key step in
the proposed framework is the construction of an FFDD dissimilarity
measure for sub-voxel alignment of fiber bundles, based on the fast marching method\index{fast marching method} (FMM). The obtained aligned WM tract-profiles enable
meaningful inter-subject comparisons and group-wise statistical analysis.
We demonstrate our method using two different datasets
of contact sports players\index{contact sports players}. Along-tract pairwise comparison as well as group-wise analysis, with respect to non-player healthy controls, reveal
significant and spatially-consistent FFDD anomalies. 
Comparing our method with along-tract FA analysis shows improved sensitivity to subtle structural anomalies in football players over standard FA measurements.

\end{abstract}

\vspace{-0.5cm}
\section{Introduction}
\vspace{-0.2cm}
WM tractography from diffusion tensor imaging (DTI) is an efficient
tool for longitudinal analysis and group studies, in particular when
standard magnetic resonance imaging (MRI) is not sufficiently sensitive to detect subtle structural
anomalies, such as in mild traumatic brain injury\index{mild traumatic brain injury} (mTBI) \cite{shenton2012review}. The fiber bundles rendered by 
tractography, in the form of streamline 3D coordinates, can be represented by
 geometrical properties as well as diffusivity measures (e.g.,
fractional anisotropy - FA, mean diffusivity - MD, axial diffusivity
- AD, radial diffusivity - RD). Nevertheless, coherent mathematical
modeling of the bundles, for {\it along-tract} pair-wise comparison and group-wise
analysis, is a challenging task. The main difficulty
is finding a common parameterization to faithfully represent the many fibers
within a single bundle, and to match different bundles.

A straight-forward parameterization considers the natural grid of the 
images. Often, voxel-based registration of the MRI volumes is
performed prior to the modeling. However, whole-brain registration
does not guarantee an optimal alignment between corresponding fiber
tracts due to large topological differences \cite{garyfallidis2015robust,o2015does,yeatman2012tract}. 
Therefore, most along-tract analysis approaches 
use arc-length (equidistant) re-parameterization prior to quantitative analysis 
\cite{colby2012along,klein2007automatic,o2009tract,stamile2016sensitive},  
and sometimes use anatomical landmarks \cite{maartensson2013spatial} 
or crop the tract edges~\cite{yeatman2012tract} to refine the alignment. 
Alternatively, tractography-based registration methods directly align sets of fibers 
based on their geometry and shape, using their streamline 3D coordinates, 
e.g., \cite{garyfallidis2015robust,o2012unbiased}. 

A different paradigm considers parameterization that is intrinsic to specific bundles.
Yushkevich et al.~\cite{yushkevich2008structure}
used a parametric medial-surface representation of thin sheet-like fiber
structures, by projecting the volumetric data into a 2-manifold. In
a similar manner, tube-like shaped fiber bundles were modeled by
their average (midline) trajectory in~\cite{chung2010cosine,corouge2004towards,garyfallidis2012quickbundles}.
A more recent method suggests using manifold learning to achieve joint parameterization 
of fiber bundles, by mapping corresponding tracts across subjects into a latent bundle core \cite{khatami2017bundlemap}.
Other approaches circumvent the parameterization problem altogether.
In \cite{durrleman2011registration}, a metric
on WM fiber bundles is defined by the path integral of the fibers
modeled as {\it currents} with an optimally constructed
vector field. This approach has been extended
in~\cite{charon2013varifold}, using {\it varifolds}.
However, \cite{charon2013varifold,durrleman2011registration} do not provide along-tract analysis.

The contribution of the proposed framework is two-fold, referring to both
fiber bundle modeling\index{fiber bundle modeling} and alignment.  Aiming to perform quantitative
along-tract analysis\index{along-tract analysis}, we introduce the concept of \textit{Fiber-Flux
Diffusion Density} (FFDD) descriptors that couple the bundle's geometry with local diffusivity
measures. This allows diffusion-related features to be accounted for, as well as structural variations along tracts, which may not be reflected by diffusion scalars alone.
Fiber bundle modeling, in the
form of tract-profiles, is obtained by application of
these descriptors along the mean trajectory of the bundle. 
Fiber tracts alignment is addressed as a curve matching
problem between the mean trajectories of the tracts. 
The proposed dissimilarity measure is based on FFDD tract profiles, thus utilizing both diffusional and geometrical information for the alignment task, rather than relying {\it exclusively} on geometrical properties (e.g., arc-length and curvature) as in classical curve matching algorithms~\cite{sebastian2003aligning,younes1998computable}, or on scalar measurements as in FA-based registration methods~\cite{andersson2007non,Smith06}.
Moreover, unlike traditional curve matching approaches~\cite{cohen1992tracking,younes1998computable}, we do
{\it not} map one curve into another. Instead, we adapt the
FMM\footnote{The FMM was proposed by Sethian~\cite{sethian1996fast} for solving boundary value problems of the Eikonal equation.} for
curve alignment~\cite{frenkel2003curve}, to symmetrically match
pairs of tracts with sub-voxel accuracy,  based on FFDD dissimilarities as an inverse speed map.
The proposed alignment framework plays a key role in the construction 
of standardized FFDD profiles that can be considered as a bundle-specific
atlas. This atlas facilitates group-wise statistical analysis for the assessment and localization of abnormalities
in WM fiber tracts.

We demonstrate the validity of our method by performing a tract-specific
longitudinal analysis of a basketball player diagnosed with occipital
mTBI and a frontal hemorrhage, having scans one week and 6 months post-injury. 
We further conduct a cross-sectional study, comparing 13 professional
American-football players with possible mTBIs, with 17 normal control (NC)
subjects. The analysis includes five major white matter tracts: the
left and right fronto-occipital fasciculus (IFOF), left and right
corticospinal tract (CST), and the forceps minor tract (FMT). Substantial
FFDD abnormalities were found in several football players compared
to controls, mostly located at the occipital part of the IFOF and
at the central part of the forceps minor. The same regions also demonstrate
statistically significant FFDD differences between the groups, indicated by
low p-values and high standard deviation (STD).  For some players, repeated scans revealed
consistent and increased FFDD anomalies with time, even after a few
weeks off-season. Results are in line with mTBI findings from
DTI~\cite{hulkower2013decade}. 
We also demonstrate that the proposed FFDD method provides improved sensitivity to
subtle structural anomalies compared to along-tract FA analysis, due to the use of additional geometric information.

The rest of the paper is organized as follows: Section 2 presents the FFDD
descriptors, followed by an introduction of the proposed framework for fiber bundles alignment\index{fiber bundles alignment} and statistical analysis. Section 3 presents experimental results for
two different datasets of contact-sport players. We conclude in section 4.

\vspace{-0.1cm}
\section{Methods}
\vspace{-0.1cm}
\subsection{Fiber-Flux Diffusion Density Descriptor}
\vspace{-0.1cm}
A fiber bundle~$\mathcal{B}$ can be thought of as a set of similar trajectories
with a common origin and destination, along which water molecules
are diffused \cite{heimer2012human}. In the spirit of this notion,
we define a local measure for quantifying the fiber-flux of $\mathcal{B}$
through a given plane $\pi$, with normal $\hat{n}_{\pi}\left(p\right)$
at point $p\in\pi$, i.e.,
\begin{equation}
\mathcal{F_{B}\mathit{\left(\pi;p\right)}=}\frac{1}{N_{p}}\stackrel[i=1]{N_{p}}{\sum}\hat{\tau}_{i}\left(x_{i}\right)\cdot\hat{n}_{\pi}\left(p\right),
\end{equation}
where $N_{p}$ is the number of intersected fibers, $\xi = \left\{ x_{i}\right\} $
is the set of intersection points between the plane and the fiber
bundle, and $\left\{ \hat{\tau}_{i}\left(x_{i}\right)\right\} $ are
the tangents of the fibers at those points. We call $\mathcal{F_{B}}\mathit{\left(\pi;p\right)}$
the fiber-flux density (FFD) of bundle $\mathcal{B}$ at point $p$.
The plane $\pi$ is oriented such that the fiber-flux is maximized,
i.e., $\hat{n}_{\pi}\left(p\right)=\arg\underset{\hat{n}_{\pi}}{\max}\,\mathcal{F_{B}}\mathit{\left(\pi;p\right)}$.
We use an iterative approach to solve this maximization problem in the
spirit of~\cite{tagliasacchi2009curve}. We further introduce diffusivity properties into our model by extending
the FFD measure. Let $\mathcal{S}\left(x_{i}\right)$ define a diffusivity
scalar of choice (FA, MD, AD, or RD), associated with the point $x_{i}$.
We define the fiber-flux diffusion density (FFDD) as follows:
\begin{equation}
\mathcal{J_{B}\mathit{\left(\pi;p\right)}=}\frac{1}{N_{p}}\stackrel[i=1]{N_{p}}{\sum}\mathcal{S}\left(x_{i}\right)\hat{\tau}_{i}\left(x_{i}\right)\cdot\hat{n}_{\pi}\left(p\right)
\end{equation}
In practice, we refer to the FFDD as a vector $\mathcal{\vec{J}}_{\mathcal{B}}\mathit{\left(p\right)}=\mathcal{\mathcal{J}_{B}}\mathit{\left(p\right)}\hat{n}\left(p\right)$
to account for the local orientation of the fiber bundle. Note that
the set of four FFDD descriptors (each assigned with a different diffusivity
measure) couples diffusion measures with local geometrical features
of the bundle. For example, local differences in orientation are taken
into account, and regions with ``incoherent'' fiber orientations
are ``punished'' by having lower FFDD values.
\subsection{Along Tract Profiles}
\vspace{-0.1cm}
We calculate the mean fiber of the bundle $c(s)=\left(x\left(s\right),y\left(s\right),z\left(s\right)\right)$,
where $s$ is its arc-length parameter, based on Fourier descriptor~\cite{chung2010cosine}. According to this method, 
individual streamline fibers are represented by the coefficients of cosine series expansions, which are
computed from tractography data using least squares estimation. The mean fiber is then optimally obtained 
by averaging the representation coefficients and applying the inverse transformation.
The locations of the planar cross-sections along the bundle are
determined by equidistant sampling points along the mean fiber $\left\{ p_{m}\right\} =\left\{ c(s^{m})\right\} _{m=1}^{M}$. 
\textit{Tract-profiles} $\left\{ \mathit{\mathcal{\vec{J}}_{\mathcal{B}}\mathit{\left(p_{m}\right)}}\right\} _{m}$
are obtained by applying the FFDD descriptors along the tract, over
these points. 
\vspace{-0.1cm}
\subsection{Fiber Bundles Alignment}
\begin{figure}[t!]
\begin{centering}
\setlength{\tabcolsep}{-13pt}%
\begin{tabular}{cccc}
\includegraphics[bb=0bp 0bp 525bp 420bp,clip,scale=0.22]{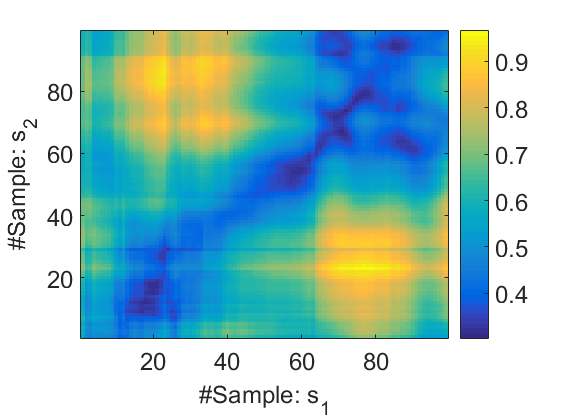} & \includegraphics[bb=0bp 0bp 525bp 420bp,clip,scale=0.22]{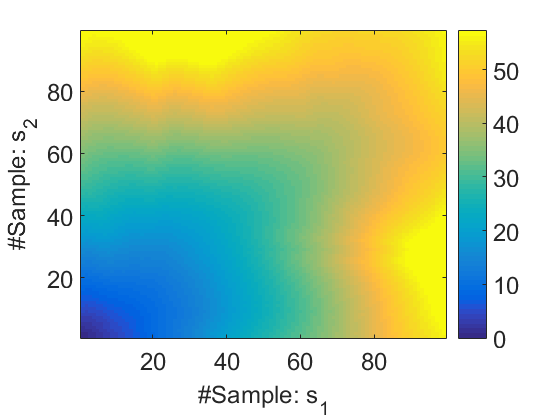} & \includegraphics[bb=0bp 0bp 525bp 420bp,clip,scale=0.22]{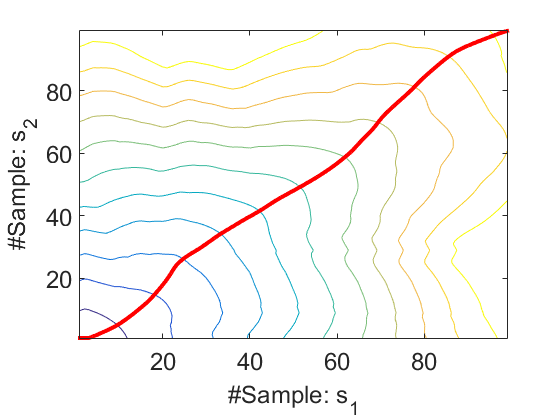} & \includegraphics[scale=0.22]{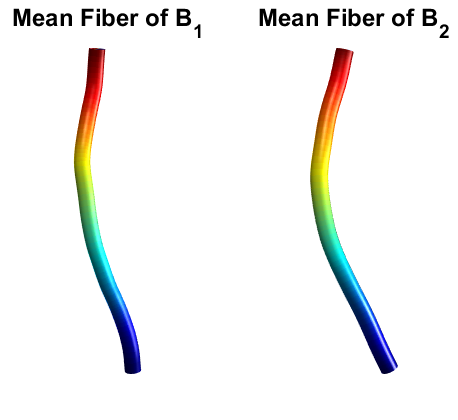}\tabularnewline
(a) $F\left(s_{1},s_{2}\right)$~~~~~~~~ & (b) $T\left(s_{1},s_{2}\right)$~~~~~~~~ & (c) Compute $\alpha$ ~~~~~~~ & (d) Aligned Curves~~~~\tabularnewline
\end{tabular}
\par\end{centering}
\caption{\small Alignment using FMM. (a) Local dissimilarities $F\left(s_{1},s_{2}\right)$
based on FFDD profiles. (b) $T\left(s_{1},s_{2}\right)$ is obtained
by solving the Eikonal equation. (c) The alignment path $\alpha\left(\tau\right)$
is computed by backtracking along the gradients of $T\left(s_{1},s_{2}\right)$.
(d) Resulting alignment (color-coded). }
\label{fig:FMM}
\vspace{-0.1cm}
\end{figure}

\vspace{-0.1cm}
We address the alignment of two bundles
$\mathcal{B}_{1}$ and $\mathcal{B}_{2}$ as a curve-matching problem 
between their 
mean fibers $c_{1}\left(s_{1}\right)$ and $c_{2}\left(s_{2}\right)$,
where $s_{1}\in\left[0\,,\,L_{1}\right]$ and
$s_{2}\in\left[0\,,\,L_{2}\right]$ are the
respective arc-length parameterizations.
We adapt the FMM-based symmetrical curve matching framework
of~\cite{frenkel2003curve} to allow sub-sampling resolution of the alignment.
Nevertheless, rather than using geometrical
properties alone for the construction of the inverse speed map $F\left(s_{1},s_{2}\right)$, 
we propose a new dissimilarity measure which relies on the FFDD profiles:
\begin{equation}
F\left(s_{1},s_{2}\right)=\left\Vert \mathcal{\vec{J}}_{\mathcal{B_{\mathrm{1}}}}\mathit{\left(s_{1}\right)}-\mathcal{\vec{J}}_{\mathcal{B_{\mathrm{2}}}}\mathit{\left(s_{2}\right)}\right\Vert +\lambda
\end{equation}
where $\lambda$ is a scalar used for regularization, set as in \cite{frenkel2003curve}.
Given  $F\left(s_{1},s_{2}\right)$, the FMM solves the  Eikonal equation $\left|\nabla
  T\left(s_{1},s_{2}\right)\right|=F\left(s_{1},s_{2}\right)\,\,\forall\,s_{1},s_{1},$
providing as output the weighted distance matrix $T\left(s_{1},s_{2}\right).$
Fig.\ref{fig:FMM}a-b present $F\left(s_{1},s_{2}\right)$ and
$T\left(s_{1},s_{2}\right)$, respectively.
The optimal alignment is then defined by 
the shortest path in $F\left(s_{1},s_{2}\right)$ from the starting
point $\left(0,0\right)$ to the endpoint $\left(L_{1},L_{2}\right)$.
The alignment path $\alpha\left(\tau\right)=\left(s_{1}\left(\tau\right),s_{2}\left(\tau\right)\right)$
defines pairs of matching points between the bundles, and is computed with sub-voxel resolution as follows:
\begin{equation}
\alpha\left(\tau-\varepsilon\right)=\alpha\left(\tau\right)-\varepsilon\nabla T\left(s_{1},s_{2}\right)\,\,;\,\,\alpha\left(\mathcal{L}\right)=\left(L_{1},L_{2}\right)
\end{equation}
as illustrated in Fig. \ref{fig:FMM}c. The step size $\varepsilon$ is usually
set to some small value ($\varepsilon\ll1).$ For uniformity,
we re-sample $\alpha$ into $M$ samples, i.e., $\left\{ \alpha\left(\tau^{m}\right)\right\} _{m=1}^{M}$,
such that the aligned mean fibers are obtained by $\tilde{C}_{1}=\left\{ c_{1}\left(s_{1}\left(\tau^{m}\right)\right)\right\} _{m=1}^{M}$
and $\tilde{C}_{2}=\left\{ c_{2}\left(s_{2}\left(\tau^{m}\right)\right)\right\} _{m=1}^{M}$
(see Fig. \ref{fig:FMM}d), and their tract-profiles are aligned accordingly:
$\tilde{J_{\mathcal{B}_{1}}}=\left\{ \mathcal{\vec{J}}_{\mathcal{B_{\mathrm{1}}}}\mathit{\left(c_{1}\left(s_{1}\left(\tau^{m}\right)\right)\right)}\right\} _{m=1}^{M}$
and $\tilde{J_{\mathcal{B}_{2}}}=\left\{ \mathcal{\vec{J}}_{\mathcal{B_{\mathrm{2}}}}\mathit{\left(c_{2}\left(s_{2}\left(\tau^{m}\right)\right)\right)}\right\} _{m=1}^{M}$.

\vspace{-0.1cm}
\subsection{Along-Tract Variability Analysis}
\vspace{-0.1cm}
\textbf{Pairwise Comparison:} Let $\tilde{J_{\mathcal{B}_{1}}}$ and
$\tilde{J_{\mathcal{B}_{2}}}$ be a pair of aligned tract-profiles
to be compared, e.g., of a subject-specific tract in two longitudinal
scans. We define a pointwise dissimilarity measure between them as
follows: 
\begin{equation}
d_{\mathcal{J}}\left(\mathcal{B}_{1},\mathcal{B}_{2};\alpha\left(\tau^{m}\right)\right)=\left\Vert \mathcal{\vec{J}}_{\mathcal{B_{\mathrm{1}}}}\mathit{\left(c_{1}\left(s_{1}\left(\tau^{m}\right)\right)\right)}-\mathcal{\vec{J}}_{\mathcal{B_{\mathrm{2}}}}\mathit{\left(c_{2}\left(s_{2}\left(\tau^{m}\right)\right)\right)}\right\Vert 
\end{equation}
Although we focus here on computing local dissimilarities along the
two bundles, global dissimilarity can also be calculated: $D_{\mathcal{J}}\left(\mathcal{B}_{1},\mathcal{B}_{2}\right)=\underset{\alpha}{\int}d_{\mathcal{J}}\left(\mathcal{B}_{1},\mathcal{B}_{2};\alpha\left(\tau\right)\right)d\alpha$.

\textbf{Group-Wise Statistical Analysis:} Alignment of multiple fiber
tracts for group-wise analysis is performed as follows. Let $\mathcal{J_{\mathrm{\mathit{g}}}=}\left\{ \mathcal{\vec{J}}_{\mathcal{B_{\mathrm{n}}}}\left(s\right)\right\} _{n=1}^{\mathcal{N}_{g}}$
denote the set of $\mathcal{N}_{g}$ tract-profiles of a group of
subjects, and let $\mathcal{C_{\mathit{g}}=}\left\{ C_{n}\left(s\right)\right\} _{n=1}^{\mathcal{N}_{g}}$
denote their respective mean fibers with a joint arc-length parameterization
$s$. We define a {\it reference} tract profile, with its corresponding
mean fiber as follows: 
\begin{equation}
\mathcal{\vec{J}}_{ref}\left(s\right)=\frac{1}{\mathcal{N}_{g}}\stackrel[n=1]{\mathcal{N}_{g}}{\sum}\mathcal{\vec{J}}_{\mathcal{B_{\mathrm{n}}}}\left(s\right)\,\,,\,\,C_{ref}\left(s\right)=\frac{1}{\mathcal{N}_{g}}\stackrel[n=1]{\mathcal{N}_{g}}{\sum}C_{n}\left(s\right)
\end{equation}
Alignment of the tract-profiles is obtained by first mapping each
of them to the reference tract as discussed in Section 2.3. We then
interpolate the resulting alignment paths $\left\{ \alpha\left(\tau_{n}\right)\right\} _{n=1}^{\mathcal{N}_{g}}$
such that they all contain the same set of $M$ samples of the reference
tract $\left\{ C_{ref}\left(s^{m}\right)\right\} _{m=1}^{M}$. We
construct a {\it bundle-specific atlas} by pointwise averaging 
the aligned tract-profiles. This atlas represents the standardized
tract-profile of the group, which is used as a benchmark for group-wise
statistical analysis. 
\vspace{-0.1cm}
\section{Experimental Results}
\vspace{-0.1cm}
We demonstrate our FFDD method on two different datasets of contact-sports players. 
Normal control (NC) group includes scans
of healthy age-matched males. Diffusion weighted images
(DWI) of all subjects were acquired on a 3T Philips Ingenia scanner
using a single-shot, spin-echo, echo-planar imaging (EPI) sequence
(TE=106 ms, TR=9000 ms, FOV = 224x224x120 mm). A total of 60 2{[}mm{]}-thick
slices were acquired with 33 different gradient directions (b=1000
s/mm$^{2}$) with a voxel resolution of 1.75$\times$1.75$\times$2
mm. Pre-Processing included rigid alignment to the SPM MNI T1-template;
motion and eddy currents correction; DTI tensor model fitting
\cite{basser1994estimation}; and Tractography of five major tracts: left and right IFOF, left
and right CST, and the FMT~\cite{yeh2013deterministic}. All performed by DSI Studio software
(\url{http://dsi-studio.labsolver.org/}). 
The tracts were delineated by placing multiple regions
of interest (ROIs) from the JHU WM atlas~\cite{mori2008stereotaxic}.
\vspace{-0.5cm}
\subsection{Longitudinal Case Study}
\begin{figure}[t!]
\begin{centering}
\setlength{\tabcolsep}{1pt}%
\begin{tabular}{cccc|cc}
\multirow{1}{*}[0.19\linewidth]{\rotatebox{90}{ONE WEEK}} & \multirow{1}{*}[0.235\columnwidth]{\includegraphics[scale=0.16]{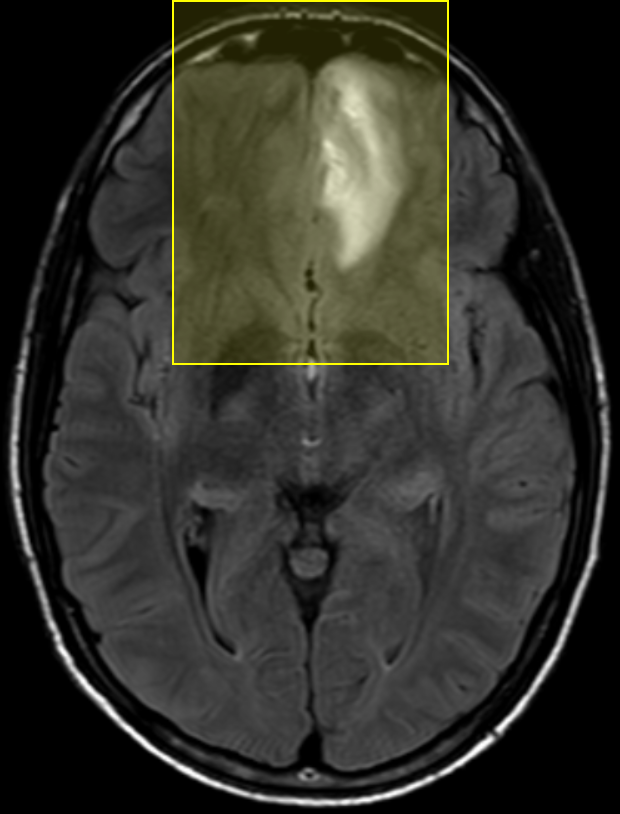}} & \multirow{1}{*}[0.235\columnwidth]{\includegraphics[scale=0.6]{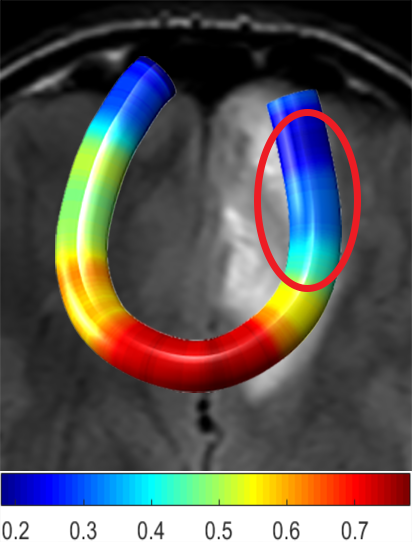}} & \multirow{1}{*}[0.235\columnwidth]{\includegraphics[scale=0.6]{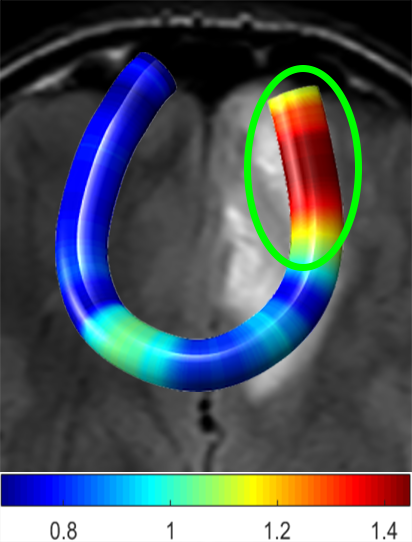}} & \includegraphics[scale=0.3]{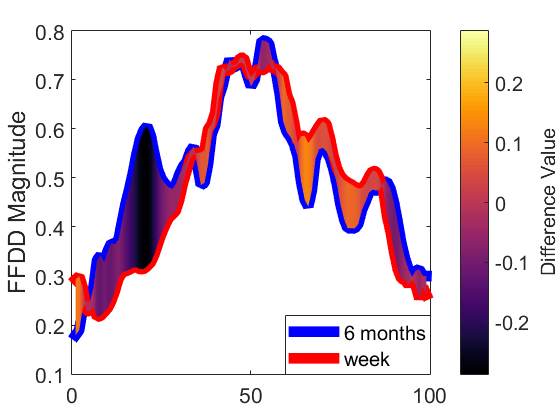} &
\multirow{1}{*}[0.2\columnwidth]{\rotatebox{270}{FA-FFDD}}\tabularnewline
\noalign{\vskip-7pt}
\multirow{1}{*}[0.19\linewidth]{\rotatebox{90}{6 MONTHS}} & \multirow{1}{*}[0.235\columnwidth]{\includegraphics[scale=0.16]{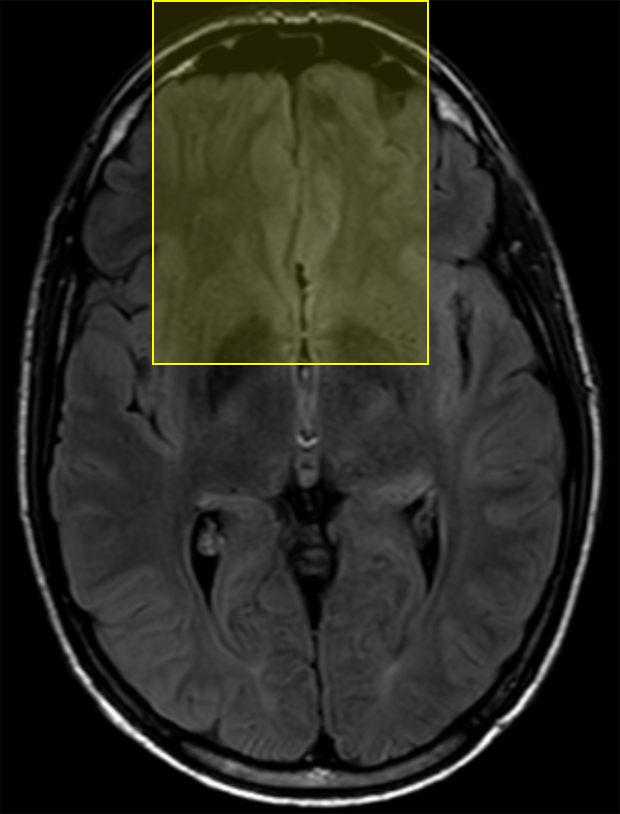}} & \multirow{1}{*}[0.235\columnwidth]{\includegraphics[scale=0.6]{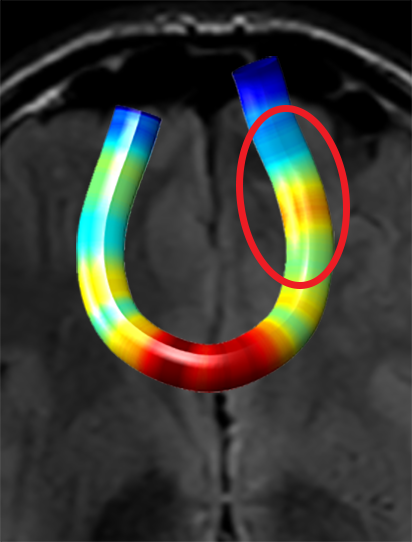}} & \multirow{1}{*}[0.235\columnwidth]{\includegraphics[scale=0.6]{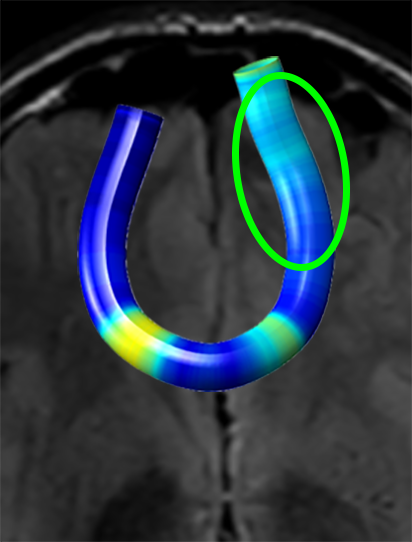}} & \includegraphics[scale=0.3]{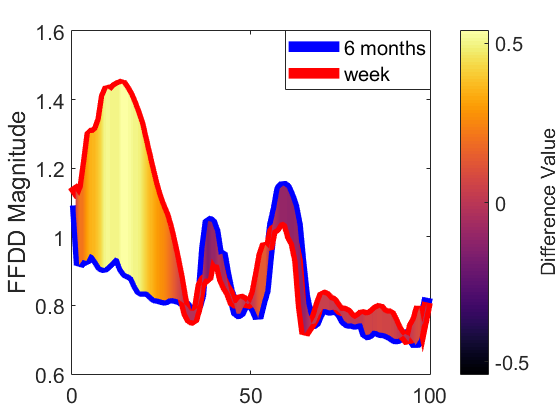} & \multirow{1}{*}[0.195\columnwidth]{\rotatebox{270}{MD-FFDD}}\tabularnewline
\noalign{\vskip-7pt}
 & (a) FLAIR & (b) FA-FFDD & \multicolumn{1}{c}{(c) MD-FFDD} & (d) Local Differences & \tabularnewline
\end{tabular}
\par\end{centering}
\centering{}\caption{\small Longitudinal FFDD analysis of the FMT. \textbf{Right Panel:} FA- and
MD-based tract-profiles of both scans. Local differences are color-coded
on the graph. \textbf{Left Panel:} FLAIR scans (axial slices), one
week post-injury (\textbf{top}) and 6 months post-injury (\textbf{bottom}).
(a) Highlighted boxes around the hemorrhaging area - lesion is no
longer visible 6 months after injury. In (b) and (c) the tracts are colored-coded 
by the magnitude of their FFDD profiles. 
Regions with high longitudinal variability
(marked in red and green ellipses) correspond to the lesion area.}
\label{fig:basketball}
\vspace{-0.1cm}
\end{figure}

\vspace{-0.1cm}
We performed pairwise comparison between two scans of a 32-year-old basketball
player, diagnosed with mild occipital traumatic brain injury and
frontal hemorrhage due to contrecoup impact, acquired one week and
6 months post-injury. The hemorrhagic lesion at the frontal right
hemisphere of the player is no longer visible in the FLAIR image acquired
6 months after injury (Fig.~\ref{fig:basketball}a). Local differences
between corresponding, longitudinal FA- and MD-FFDD profiles of the
FMT (chosen due to its proximity to the lesion area) are shown in
Fig.~\ref{fig:basketball}d.
Figs.~\ref{fig:basketball}b-c present color-coded FMT to visually
demonstrate these differences.
Results show significant longitudinal variability at the right hemisphere part of the tract, corresponding to the lesion area, and relatively minor differences along the rest of the tract. These results should be considered as a proof of concept, validating the FFDD analysis results for the detection and localization of mTBI-related variabilities between fiber bundles.

\vspace{-0.1cm}
\subsection{Football Players Study}
\begin{figure}[t!]
	\setlength{\tabcolsep}{-2pt}%
	\begin{tabular}{ccccccc}
		& Left IFOF & Right IFOF & Left CST & Right CST & Forceps Minor & \tabularnewline
		\multirow{1}{*}[0.07\linewidth]{\rotatebox{90}{Tract}~~} & \includegraphics[scale=0.15]{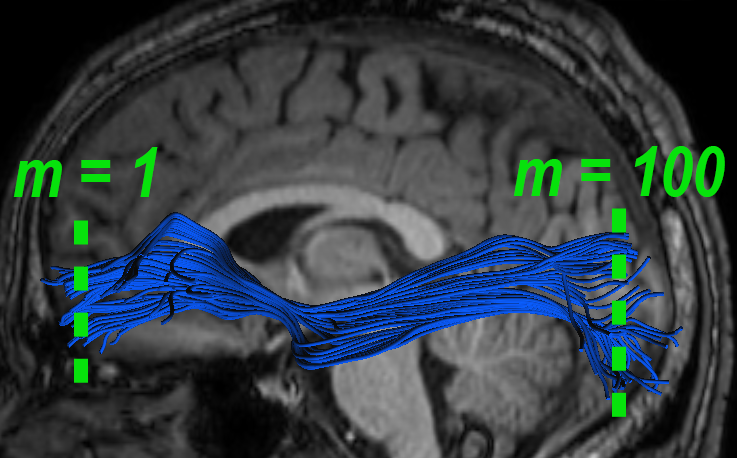} & 
		\includegraphics[scale=0.15]{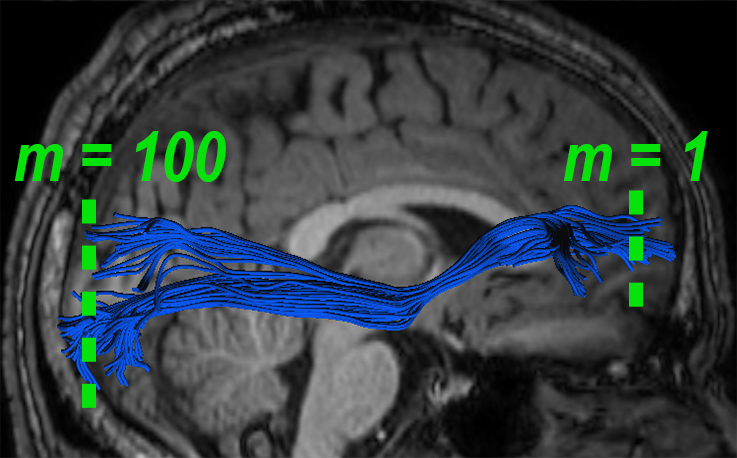} & 
		\includegraphics[scale=0.15]{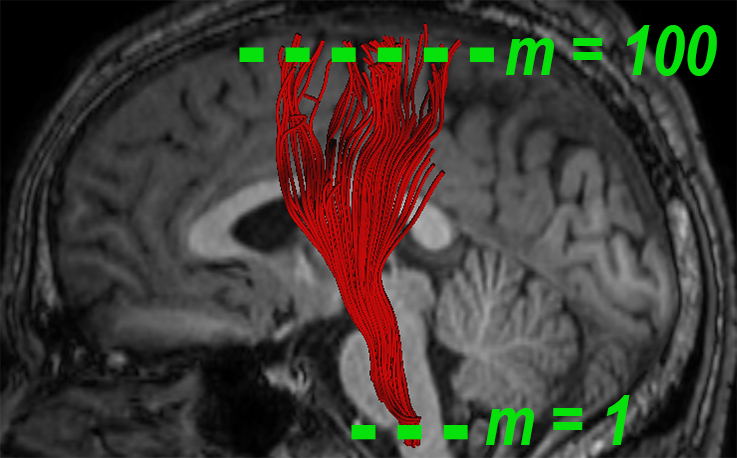} & 
		\includegraphics[scale=0.15]{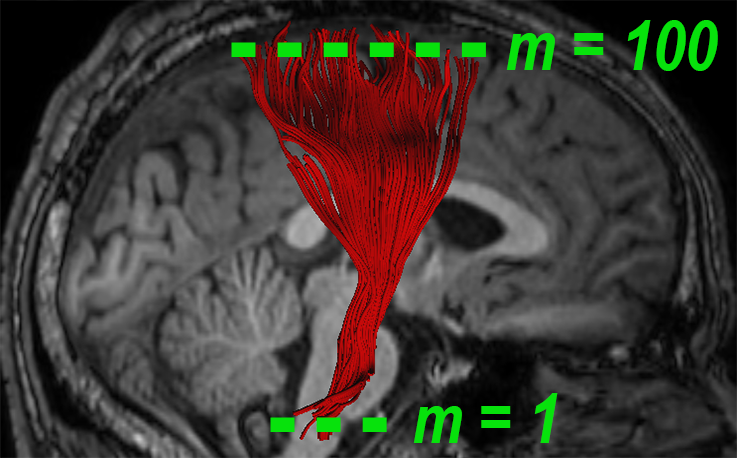} & 
		\includegraphics[scale=0.15]{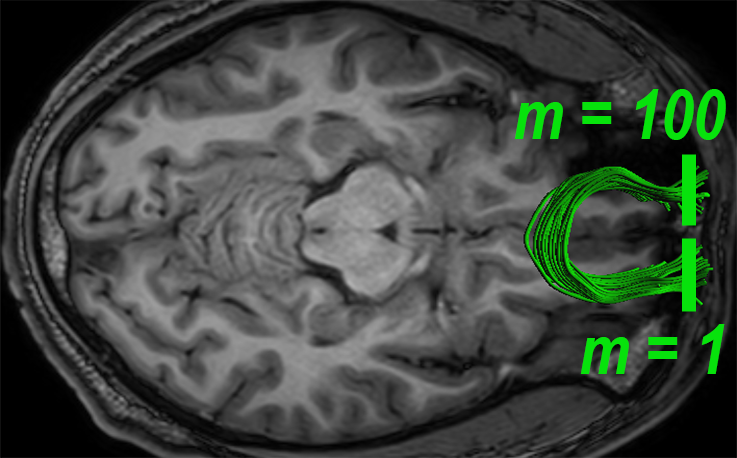} & \tabularnewline
		
		
		\multirow{1}{*}[0.125\linewidth]{\rotatebox{90}{FA-FFDD}~~} &
		\includegraphics[scale=0.17]{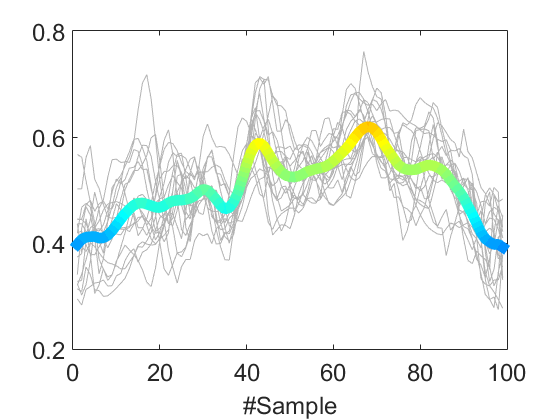} & 
		\includegraphics[scale=0.17]{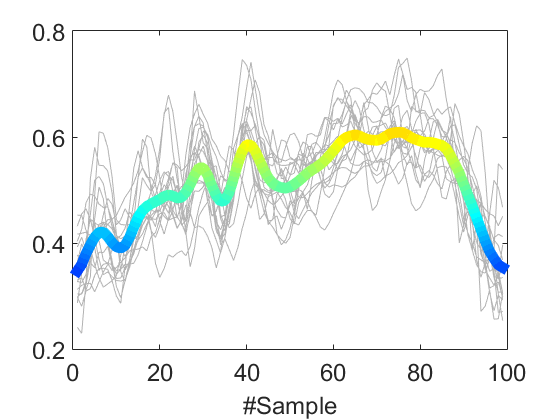} & \includegraphics[scale=0.17]{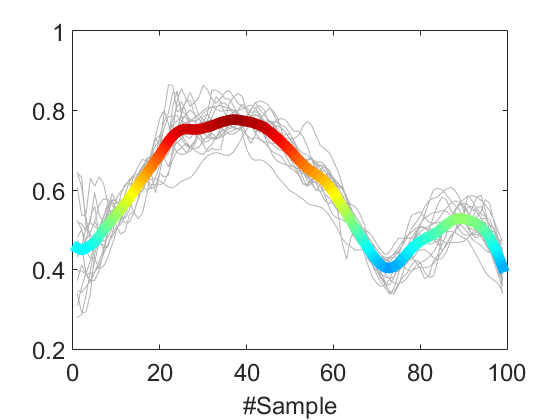} & 
		\includegraphics[scale=0.17]{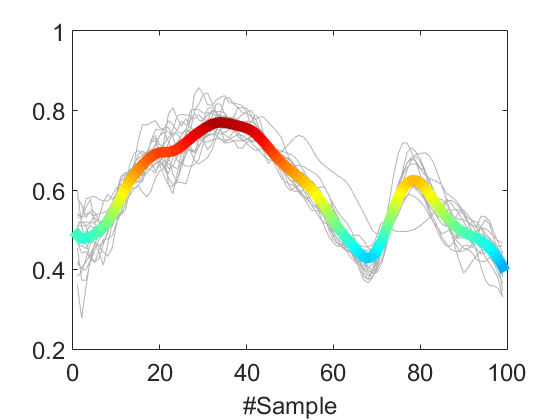} & \includegraphics[scale=0.17]{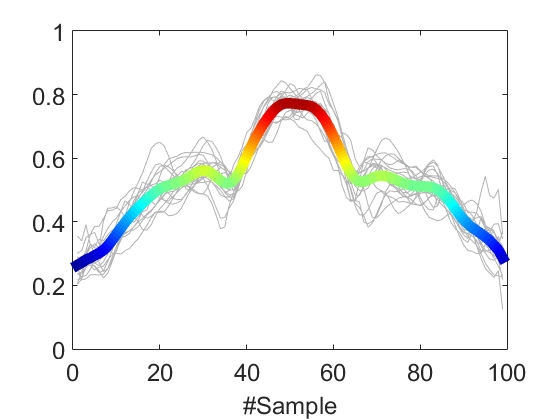} & \multirow{1}{*}[0.126\columnwidth]{\includegraphics[scale=0.16]{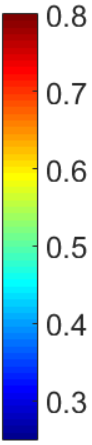}}\tabularnewline
		
		\multirow{1}{*}[0.125\linewidth]{\rotatebox{90}{MD-FFDD}~~} &
		\includegraphics[scale=0.17]{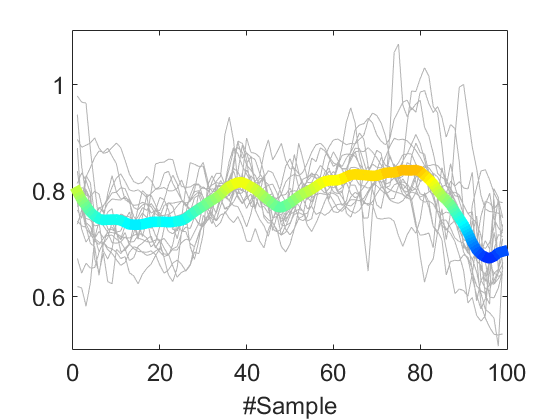} & 
		\includegraphics[scale=0.17]{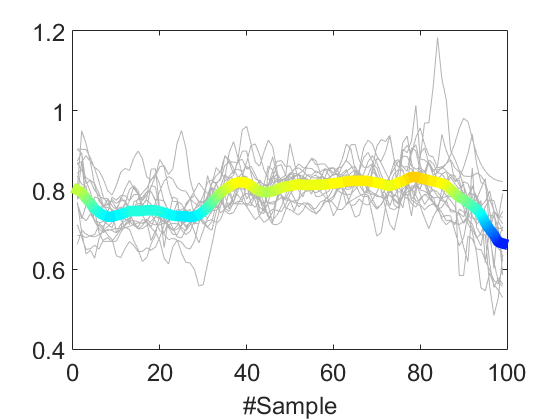} & 
		\includegraphics[scale=0.17]{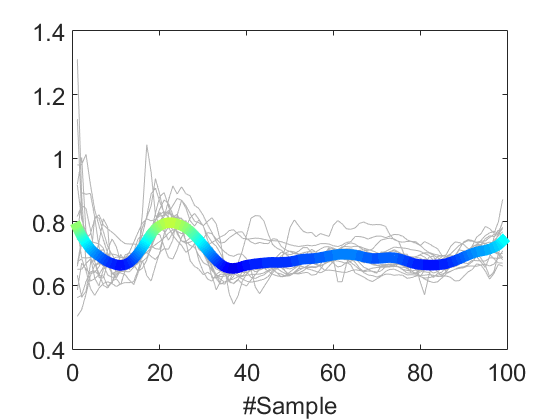} & 
		\includegraphics[scale=0.17]{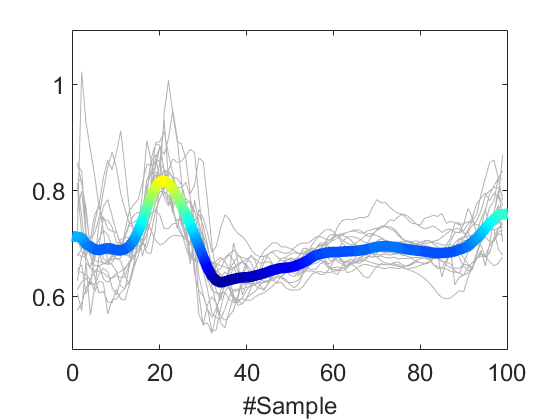} & 
		\includegraphics[scale=0.17]{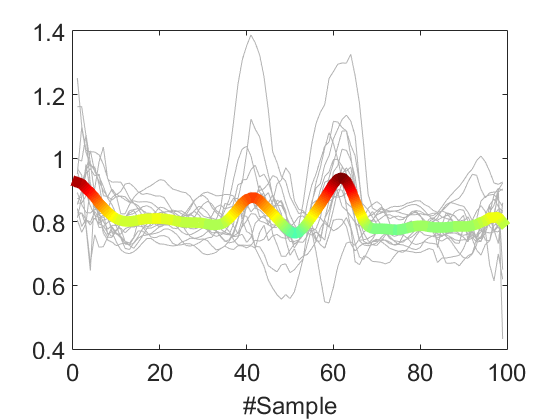} & \multirow{1}{*}[0.126\columnwidth]{\includegraphics[scale=0.16]{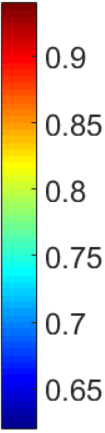}}\tabularnewline
		
		\multirow{1}{*}[0.125\linewidth]{\rotatebox{90}{AD-FFDD}~~} &
		\includegraphics[scale=0.17]{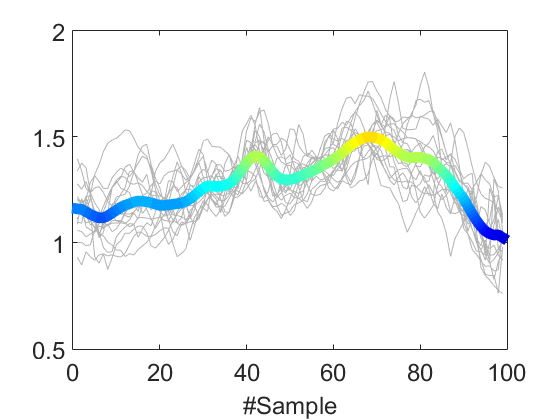} & 
		\includegraphics[scale=0.17]{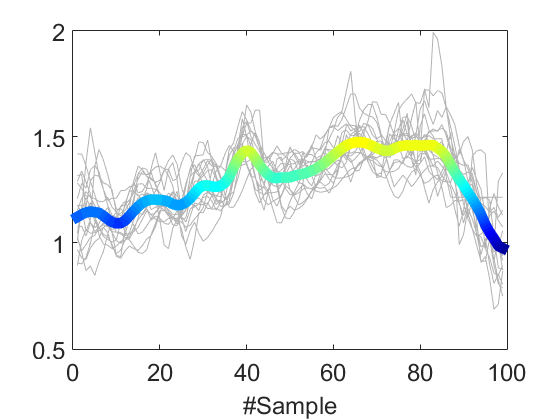} & 
		\includegraphics[scale=0.17]{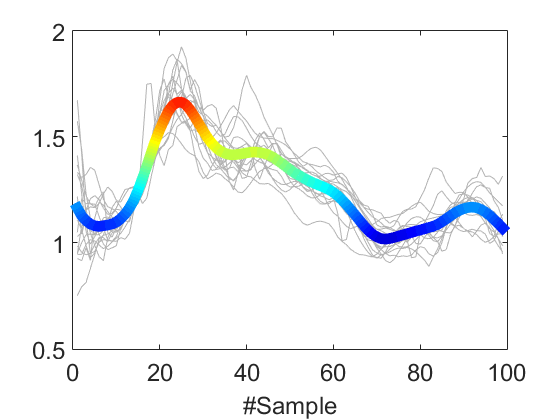} & 
		\includegraphics[scale=0.17]{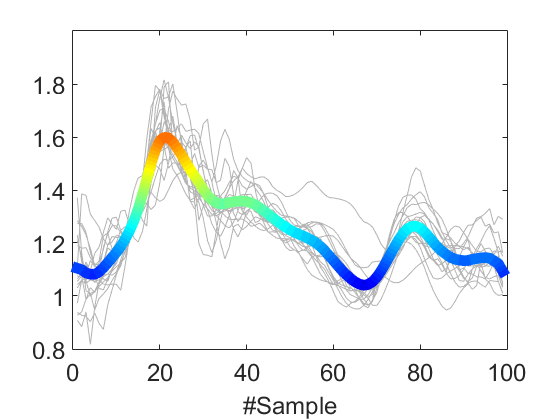} & 
		\includegraphics[scale=0.17]{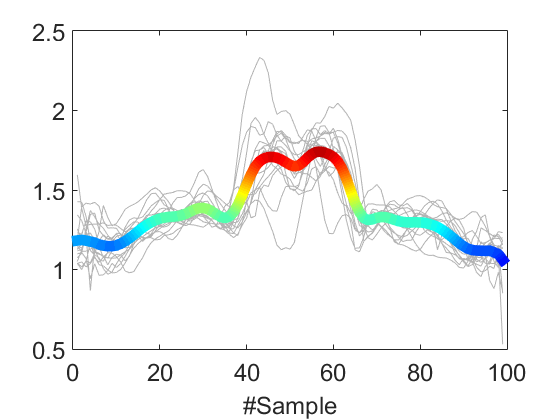} & \multirow{1}{*}[0.126\columnwidth]{\includegraphics[scale=0.16]{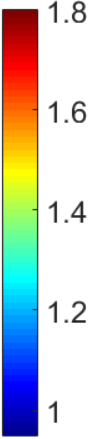}}\tabularnewline
		
		\multirow{1}{*}[0.125\linewidth]{\rotatebox{90}{RD-FFDD}~~} & \includegraphics[scale=0.17]{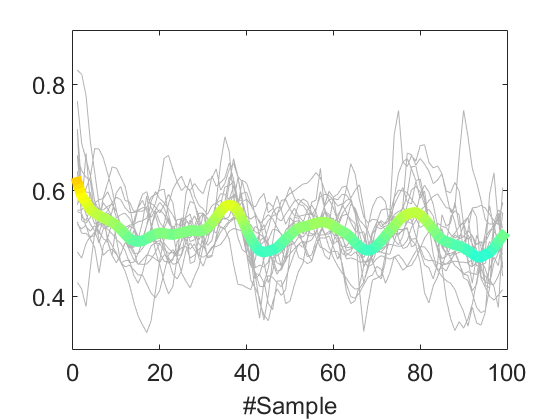} & 
		\includegraphics[scale=0.17]{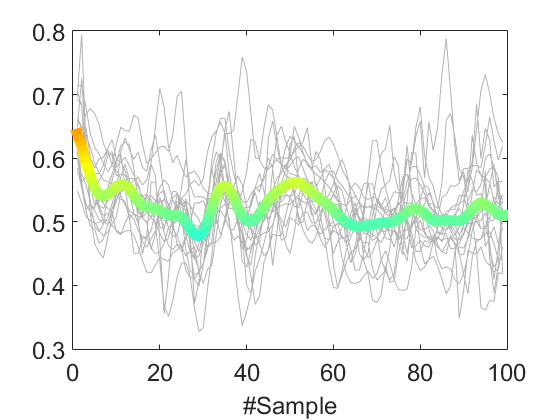} & 
		\includegraphics[scale=0.17]{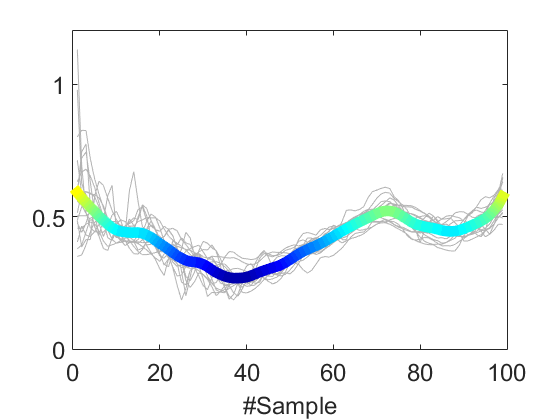} & 
		\includegraphics[scale=0.17]{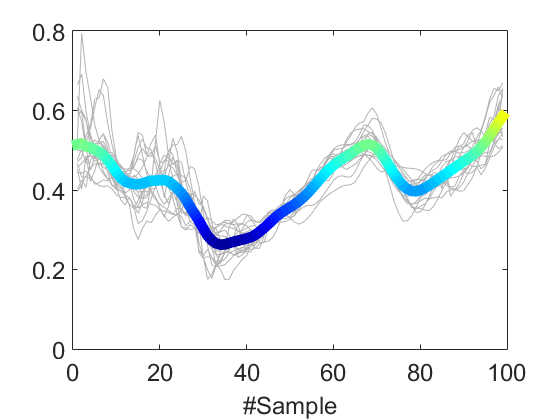} & 
		\includegraphics[scale=0.17]{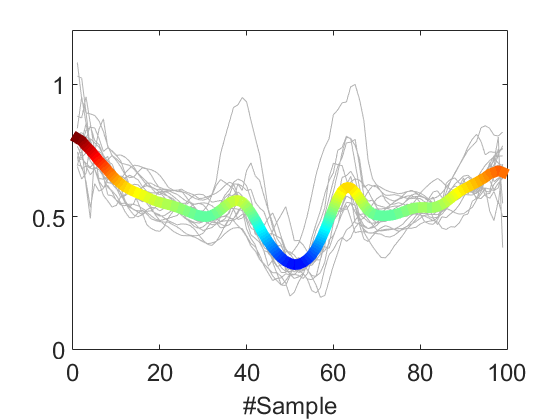} & \multirow{1}{*}[0.126\columnwidth]{\includegraphics[scale=0.16]{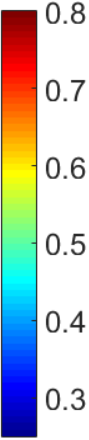}}\tabularnewline

	\end{tabular}
	
	\caption{\small \textbf{Top row:} The 5 examined tracts. \textbf{Bottom rows:} Four types of standardized tract-profiles (colored): FA-FFDD, MD-FFDD, AD-FFDD, and RD-FFDD constructed from the
	aligned tract-profiles of NCs (gray).}
	\label{fig:standardProfiles}
	\vspace{-0.1cm}
\end{figure}

\vspace{-0.1cm}
We analyzed 13 active professional American-football players (mean age = 28.3,
STD = 6.4), with respect to 17 NCs (mean age = 26.1,
STD = 2.3). For each subject, four FFDD tract-profiles were computed
(based on FA, MD, RD, and AD), for each of the five examined tracts.
The standardized FFDD profiles of NCs are shown in Fig. \ref{fig:standardProfiles}. Note that although FFDD values vary along the tracts, their
profiles are consistent across subjects.

\begin{figure}[t!]
	\begin{centering}
		\setlength{\tabcolsep}{-2.5pt}%
		\begin{tabular}{ccccc}
			& ~~Pointwise Mean & $\mathrm{\Delta}$Mean~~ & ~~Pointwise STD & $\mathrm{\Delta}$STD~~\tabularnewline
			\multirow{1}{*}[0.1\linewidth]{\rotatebox{90}{FMT}~~} & \includegraphics[scale=0.2]{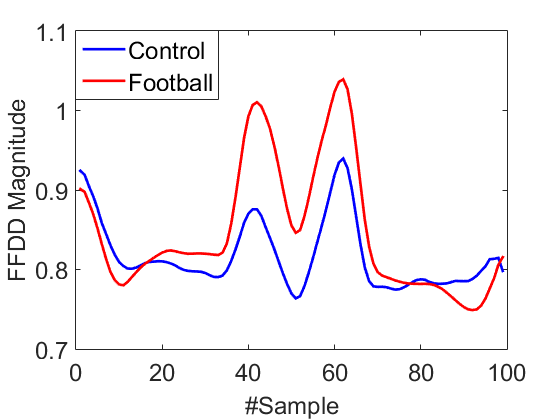} & ~~\includegraphics[scale=0.15]{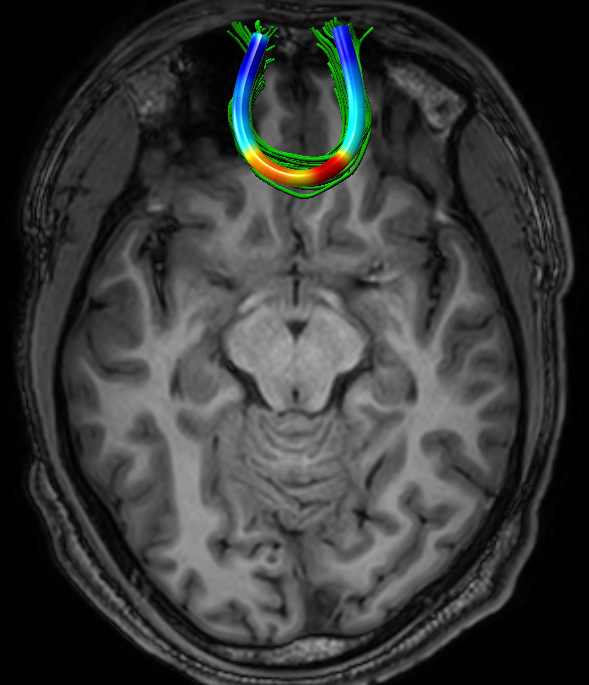}~\includegraphics[scale=0.23]{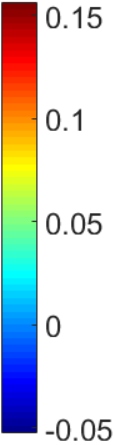} & ~\includegraphics[scale=0.2]{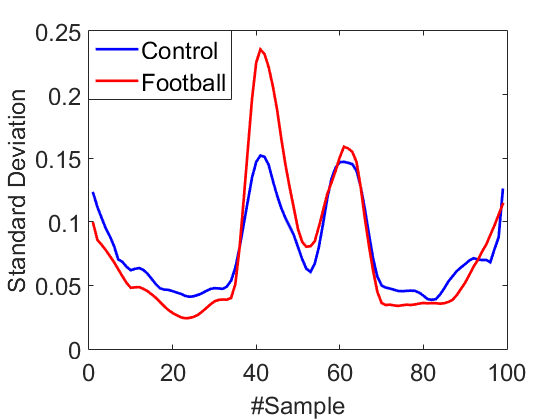} & ~~\includegraphics[scale=0.15]{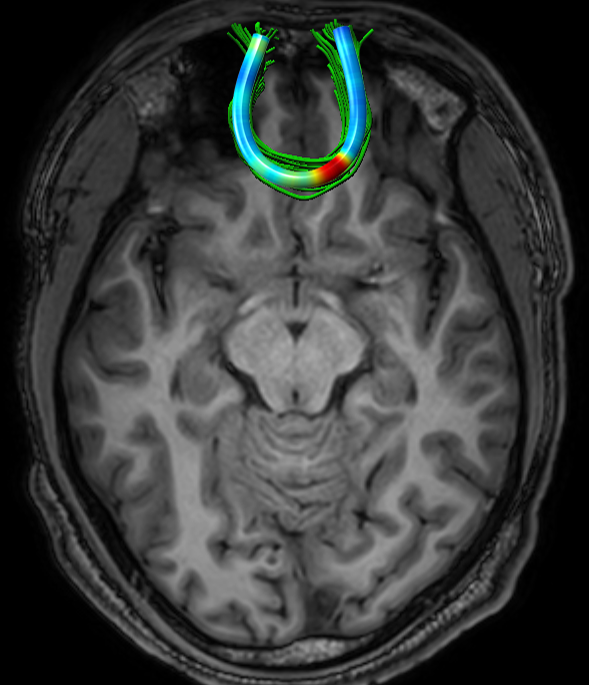}~\includegraphics[scale=0.235]{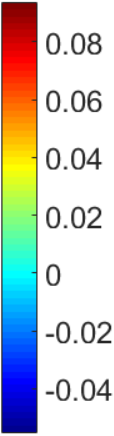}\tabularnewline[-4pt]
			\multirow{1}{*}[0.14\columnwidth]{\rotatebox{90}{Left IFOF}~~} & \includegraphics[scale=0.2]{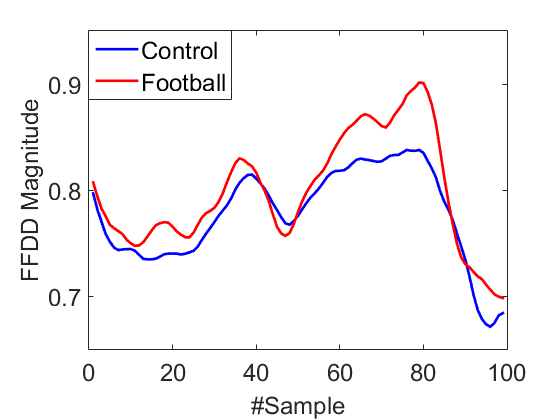} & \multirow{1}{*}[0.147\columnwidth]{\includegraphics[scale=0.172]{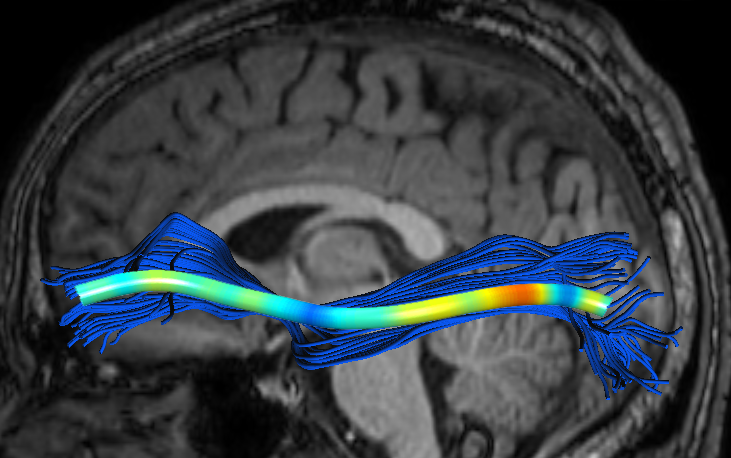}~~} & ~\includegraphics[scale=0.2]{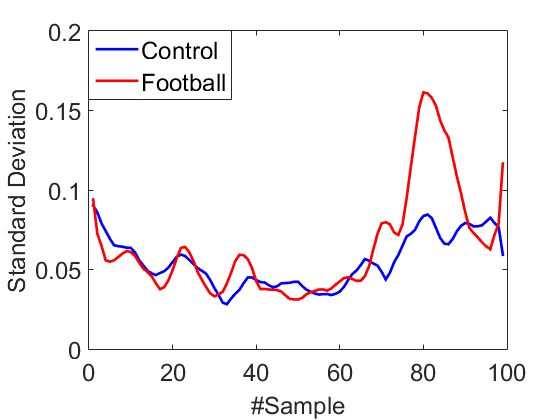} & \multirow{1}{*}[0.147\columnwidth]{\includegraphics[scale=0.172]{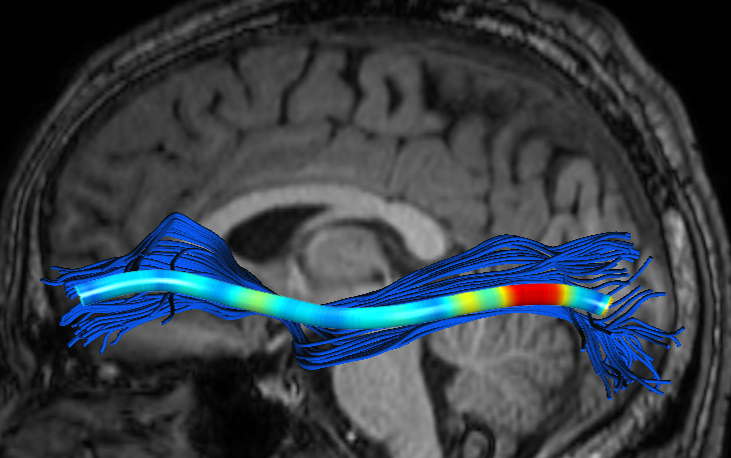}}\tabularnewline
		\end{tabular}
		\par\end{centering}
	\caption{\small Group-wise statistical analysis based on MD-FFDD. Pointwise
		comparison of within-group average profile (left) and STDs (right) are
		presented for the FMT (top) and the left IFOF (bottom).}
	\label{fig:GroupStats}
	\vspace{-0.1cm}
\end{figure}

\begin{figure}[t!]
	\centering{}\setlength{\tabcolsep}{-1pt}%
	\begin{tabular}{cc|c}
		Left IFOF & Forceps Minor & Tracts-Average Analysis\tabularnewline
		\includegraphics[scale=0.27]{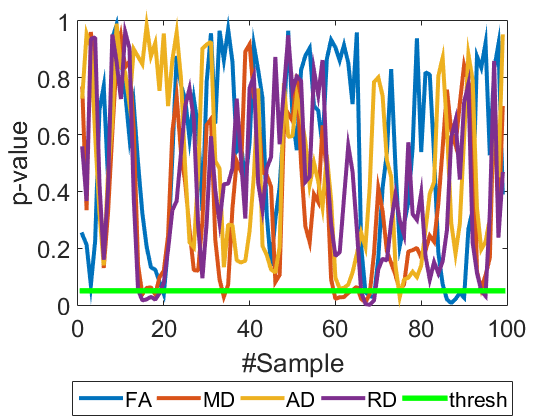} & 
		\includegraphics[scale=0.27]{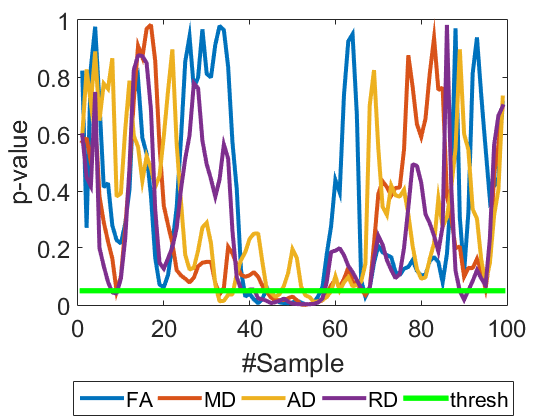} & ~\includegraphics[scale=0.27]{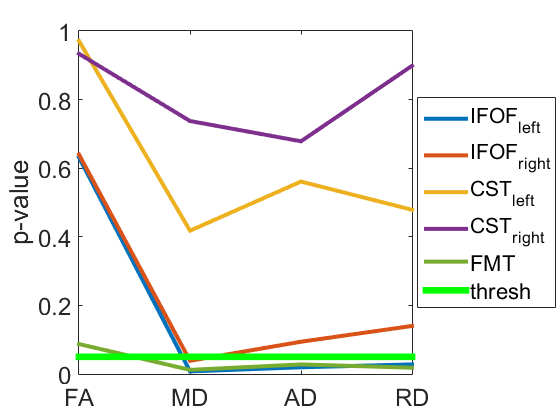}\tabularnewline
	\end{tabular}\caption{Group-wise p-values analysis. 
		\textbf{Left and middle:} pointwise (corrected) p-values along the left IFOF and FMT, based on FFDD profiles. Most statistically significant differences between the groups (p-value\textless{}0.05)
		are located in the occipital part of the IFOF (samples \#60 to \#90) 
		and central part of the FMT (samples \#40 to \#60). \textbf{Right:} scalar p-values based on tract-average diffusion measures (FA, MD, AD, and RD) of five different tracts. Statistically significant differences between the groups are shown for the left IFOF and FMT,
		in MD, AD, and RD measures.\label{fig:p-values}}
\end{figure}

\begin{figure}[t!]
	\centering{}\setlength{\tabcolsep}{-1pt}%
	\begin{tabular}{cccc}
		& MD Analysis & AD Analysis & RD Analysis\tabularnewline
		\multirow{1}{*}[0.16\columnwidth]{\begin{turn}{90}
				Left IFOF
		\end{turn}} & ~~\includegraphics[scale=0.25]{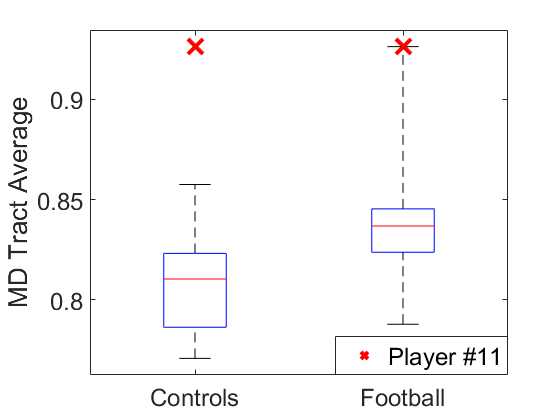} & \includegraphics[scale=0.25]{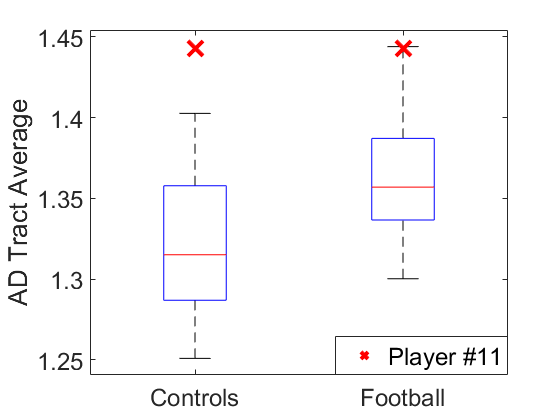} & \includegraphics[scale=0.25]{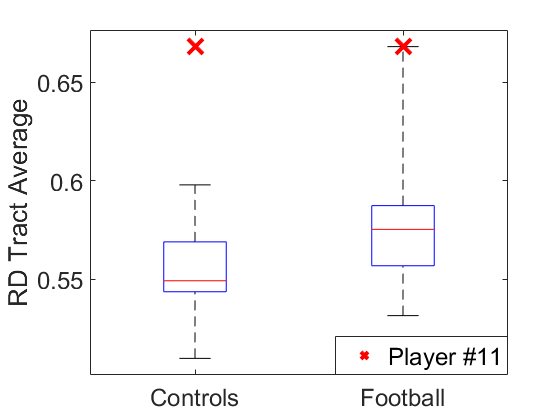}\tabularnewline
		\multirow{1}{*}[0.185\columnwidth]{\begin{turn}{90}
				Forceps Minor
		\end{turn}} & ~~\includegraphics[scale=0.25]{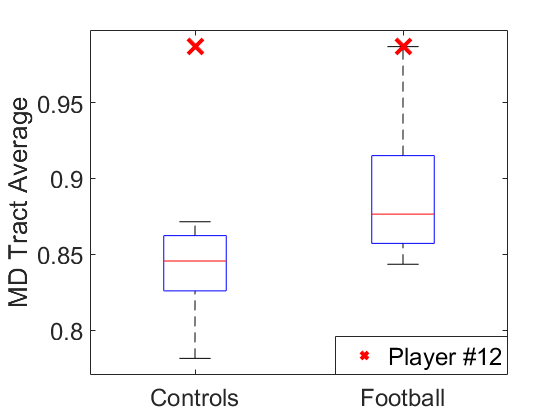} & \includegraphics[scale=0.25]{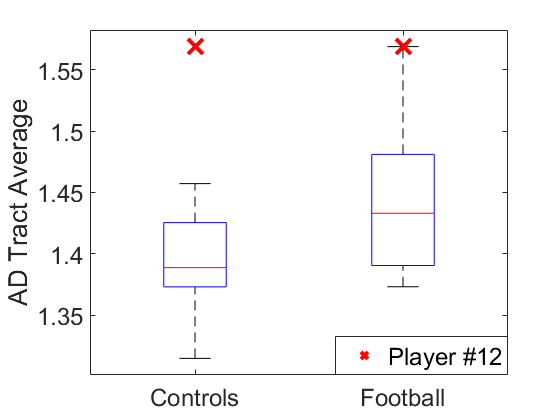} & \includegraphics[scale=0.25]{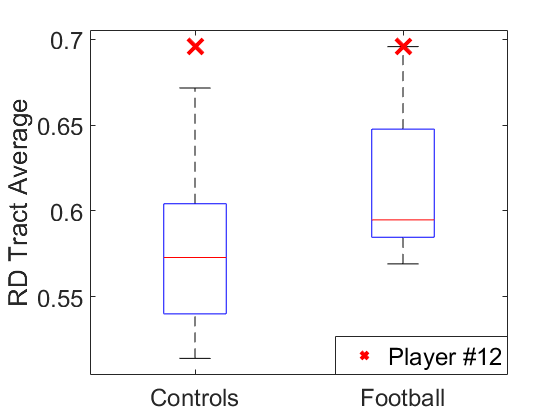}\tabularnewline
	\end{tabular}\caption{Box-plot group-wise statistical analysis of
		average diffusivity measures in the left IFOF
		(top) and FMT (bottom). Each box represents the distribution of values
		within the group: the red line represents the group-average value, the edges
		of the box represent the 25th and 75th precentiles of the group, and the
		edges of the dashed line represent the maximal and minimal values. 
		The \textbf{\textcolor{red}{x}} symbol represents the maximal value measured in the football 
		group. Note that for both tracts, the football group consistently demonstrates higher 
		values across all diffusivity measures.\label{fig:tract_average}}
		\vspace{-0.5cm}
\end{figure}

Fig. \ref{fig:GroupStats} presents pointwise group-average and STD
of MD-FFDD profiles of football players, demonstrating increased
values at the occipital part of the left IFOF, and at the central part
of the FMT, compared to NCs. 
Note that the football group also exhibits higher STD values compared to NCs, at the same areas along the tracts with increased group-average values. 
This statistical spread indicates that only a subset of the football players group has abnormal FFDD values, as expected. 
Our method also demonstrates statistically significant differences between the  
football and control groups (p-values < 0.05) at these regions of the IFOF and FMT, as shown in 
Fig.~\ref{fig:p-values}.
The left panel of the figure presents an \textit{along-tract} p-values analysis of the two tracts, 
calculated pointwise based on four different FFDD tract profiles and corrected for multiple comparisons using false detection rate (FDR) \cite{benjamini1995controlling}. 
As reference, the right panel of the figure presents a  
p-values analysis based on \textit{whole tract average} of conventional diffusivity measures
extracted via DSI Studio, calculated using an unpaired T-test, which also shows 
statistically significant differences between the groups in the IFOF and FMT for some diffusion measures. 
These findings are further supported by a group-wise statistical analysis (mean and STD) 
of \textit{whole-tract} average diffusivity measures (MD, AD, and RD), presented in Fig.~\ref{fig:tract_average}. 
Results are in line with the FFDD analysis, demonstrating increased group-average diffusivity in the left IFOF and FMT of the football group compared to NCs.
The figure also indicates the  maximal value measured within 
the football group for each diffusivity measure. 
Note that for the left IFOF, player \#11 demonstrates 
maximal values across all diffusivity measures, while the same applies 
to the FMT of player \#12. 
We note that the CST did not present significant differences between the
groups, in both FFDD and conventional analysis.

\begin{figure}[t!]
\begin{centering}
\setlength{\tabcolsep}{-1pt}%
\begin{tabular}{ccccc}
Mid vs. Post & Mid-season & Post-season &  & \tabularnewline[-3pt]
\includegraphics[scale=0.25]{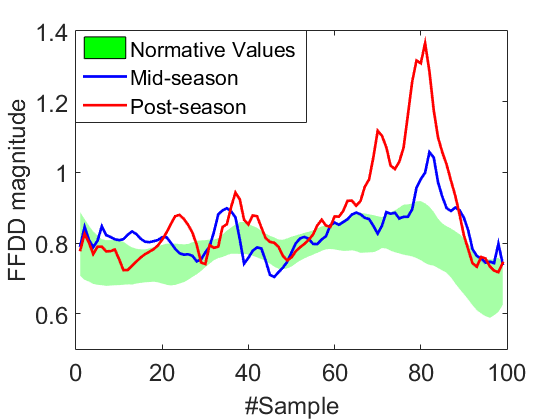} & \multirow{1}{*}[0.189\columnwidth]{\includegraphics[scale=0.19]{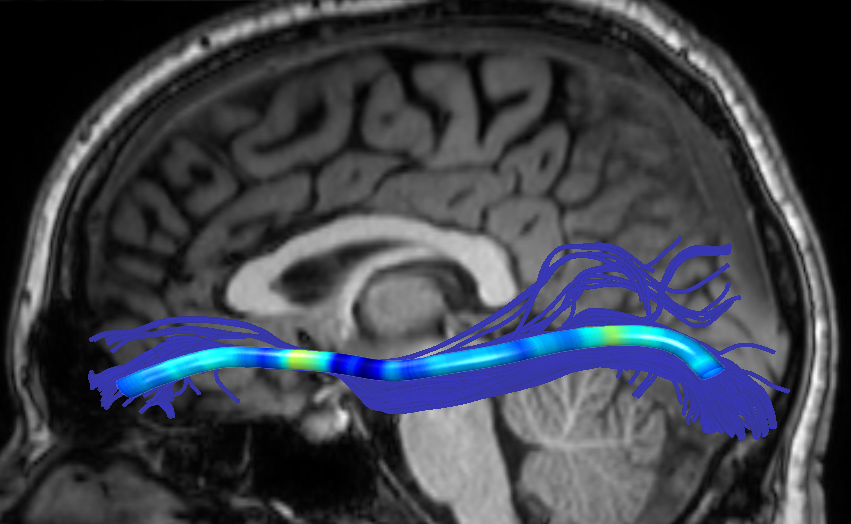}} & \multirow{1}{*}[0.189\columnwidth]{~~\includegraphics[scale=0.19]{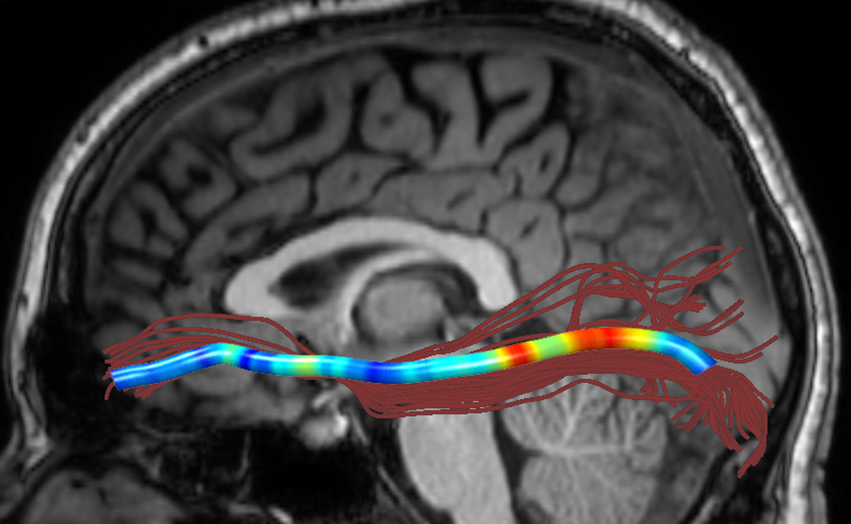}} & \multirow{1}{*}[0.19\columnwidth]{~~\includegraphics[scale=0.235]{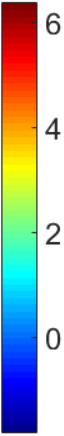}} & \multirow{1}{*}[0.16\columnwidth]{~~\rotatebox{270}{STDs}}\tabularnewline
\end{tabular}
\par\end{centering}
\caption{\small MD-FFDD longitudinal changes in the left IFOF of player \#11. \textbf{Left:}
Mid-season and post-season tract-profiles in comparison to normative
values ($\pm1$ STD from standardized profile of NC).
\textbf{Middle and Right:} Mid- and post-season abnormalities are
color-coded along the tract (in units of \#STDs from standardized
profile). In the occipital area, mid-season profile demonstrates moderate
abnormality (up to 2.5 STDs) while in post-season substantial abnormality
(up to 6.5 STDs) is shown.}
\label{fig:MidVsPost}
\end{figure}

\begin{figure}[t!]
	\centering{}\setlength{\tabcolsep}{-4.2pt}%
	\begin{tabular}{ccccc}
		Mid vs. Post & Mid-season~~~ & Post-season &  & \tabularnewline[-3pt]
		\includegraphics[scale=0.375]{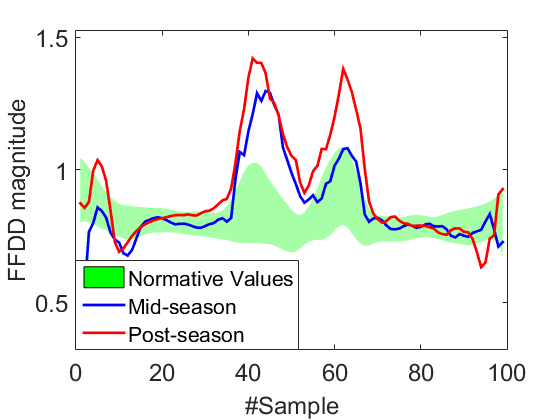} & \multirow{1}{*}[3.6cm]{\includegraphics[scale=0.181]{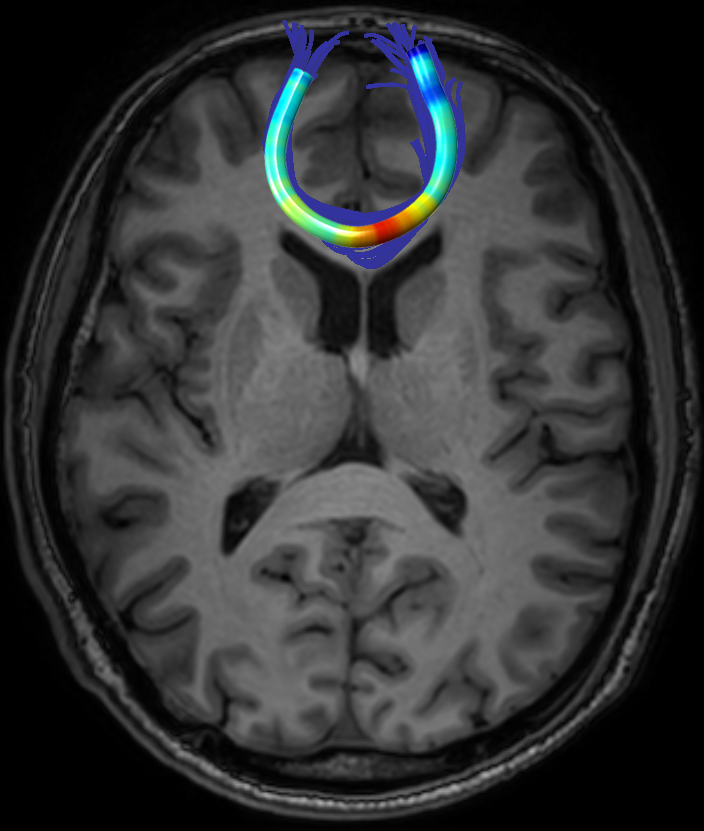}~~~~} & \multirow{1}{*}[3.6cm]{\includegraphics[scale=0.183]{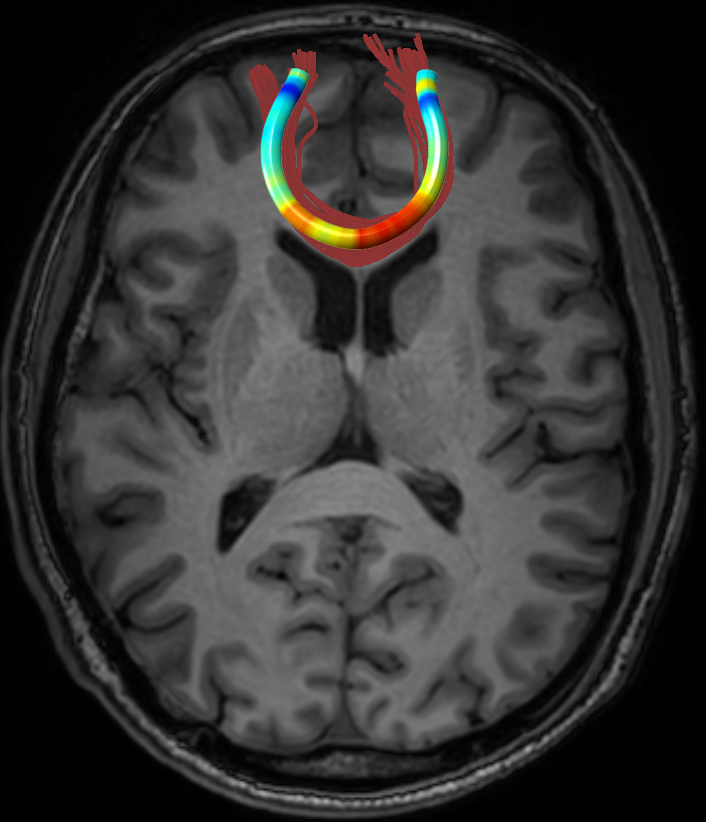}} & \multirow{1}{*}[3.6cm]{~~~~\includegraphics[scale=0.35]{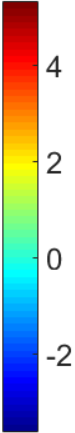}~~~~} & \multirow{1}{*}[3.2cm]{\begin{turn}{-90}
				~~~~~~STDs
		\end{turn}}\tabularnewline
	\end{tabular}\caption{MD-FFDD longitudinal changes along the FMT of player \#12. \textbf{Left:}
		Mid-season and post-season tract-profiles are compared to normative
		values ($\pm1$ STD from standardized profile of NCs).
		\textbf{Middle and Right:} Mid- and post-season abnormalities are color-coded
		along the tract (in units of \#STDs from standardized profile). While
		mid-season profile demonstrates abnormality (up to 3.5 STDs)
		in a small region around the center of the tract, the post-season profile 
		shows expansion in both magnitude (up to 4.5 STDs) and location of abnormality
		along the tract.\label{fig:MidPost_12}}
\end{figure}

Experiments also showed significant FFDD longitudinal changes between mid-season and 
post-season scans in some football players. 
Fig. \ref{fig:MidVsPost} presents mid- and post-season MD-FFDD
profiles comparison of the left IFOF of one of the players, showing
increased irregularities over time at the occipital part of the tract.
Fig. \ref{fig:MidPost_12} presents a similar MD-FFDD longitudinal analysis of the FMT of
a different player, showing increased irregularities at the central part of the tract.

\vspace{-0.1cm}
\subsection{Sensitivity Analysis}
\begin{figure}[t!]
	\begin{centering}
		\setlength{\tabcolsep}{-1pt}
		\par\end{centering}
	\begin{centering}
		\begin{tabular}{c>{\centering}p{2.7cm}>{\raggedright}m{3cm}|>{\centering}p{2.7cm}>{\raggedright}p{3cm}cl}
			& \multicolumn{2}{c|}{Left IFOF} & \multicolumn{2}{c}{Right IFOF} &  & \tabularnewline
			\cline{2-5} 
			 & {\tiny{}(a) Football vs. NCs} & {\tiny{}(b) Variability from NCs} & {\tiny{}(a) Football vs. NCs} & {\tiny{}(b) Variability from NCs} &  & \tabularnewline
			\rotatebox{90}{~~~~~FA-FFDD} & ~\includegraphics[width=2.8cm,height=2.2cm]{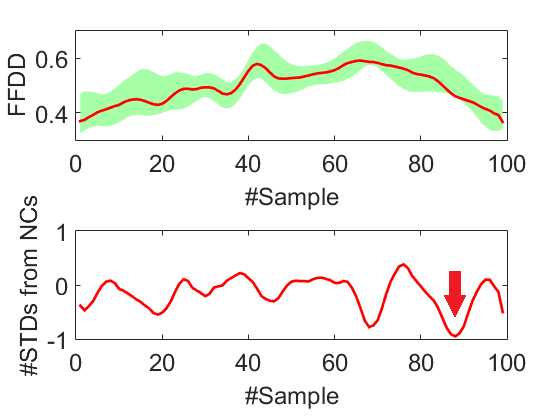} & \multirow{1}{3cm}[1.8cm]{\includegraphics[width=3cm,height=1.9cm]{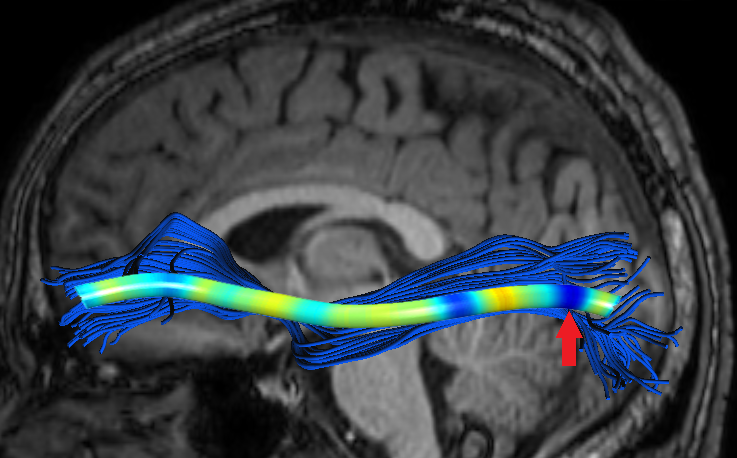}} & ~\includegraphics[width=2.8cm,height=2.2cm]{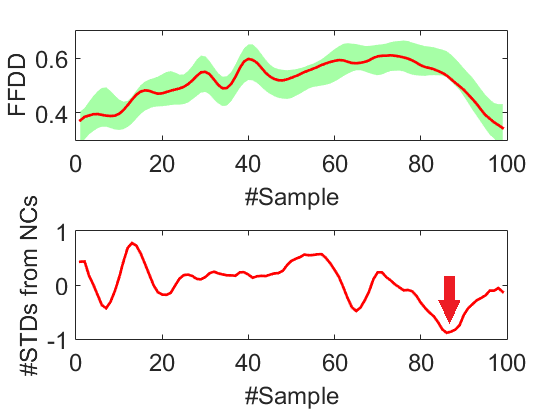} & \multirow{1}{3cm}[1.8cm]{\includegraphics[width=3cm,height=1.9cm]{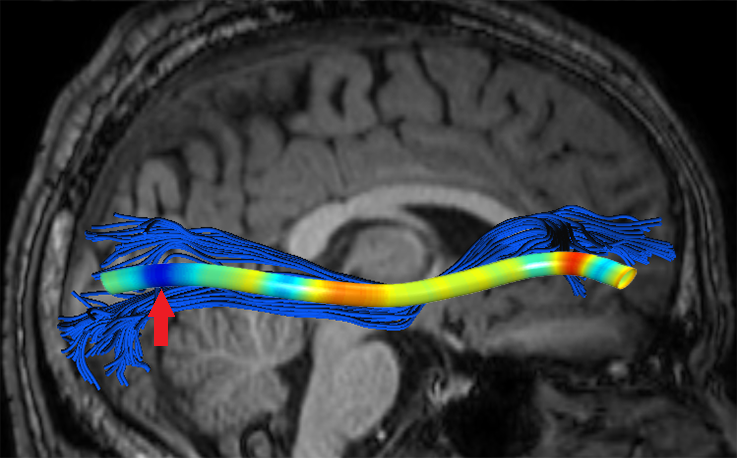}} & \multirow{2}{*}[0.15\columnwidth]{\includegraphics[width=0.7cm,height=4.25cm]{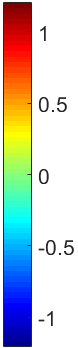}} & \multirow{2}{*}[0.3cm]{\rotatebox{270}{STDs}}\tabularnewline
			\rotatebox{90}{~~Standard FA} & ~\includegraphics[width=2.8cm,height=2.2cm]{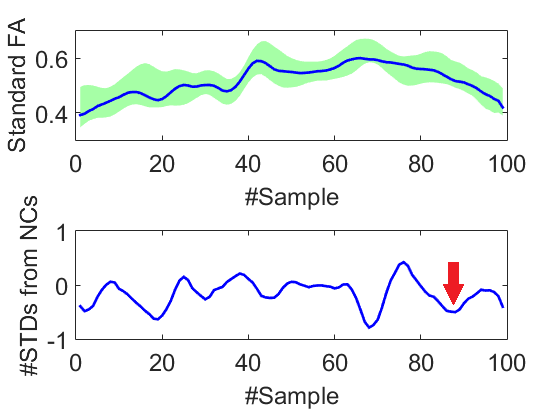} & \multirow{1}{3cm}[1.8cm]{\includegraphics[width=3cm,height=1.9cm]{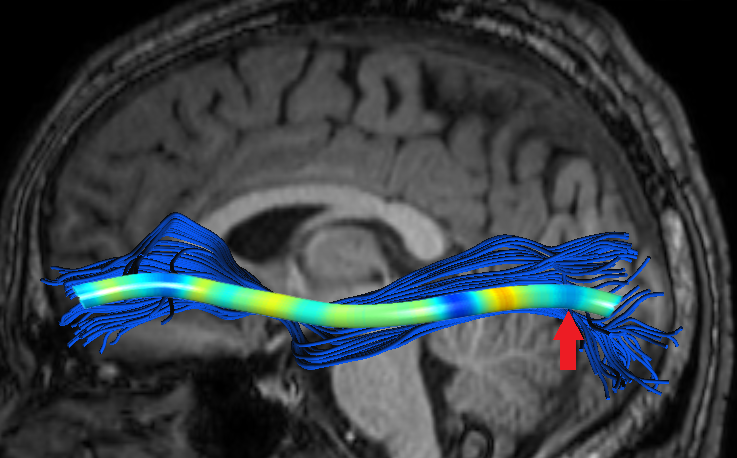}} & ~\includegraphics[width=2.8cm,height=2.2cm]{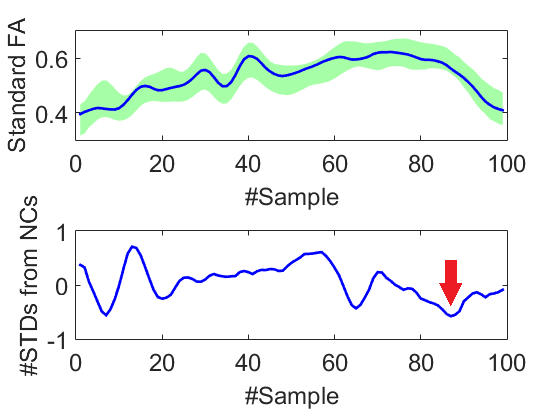} & \multirow{1}{3cm}[1.8cm]{\includegraphics[width=3cm,height=1.9cm]{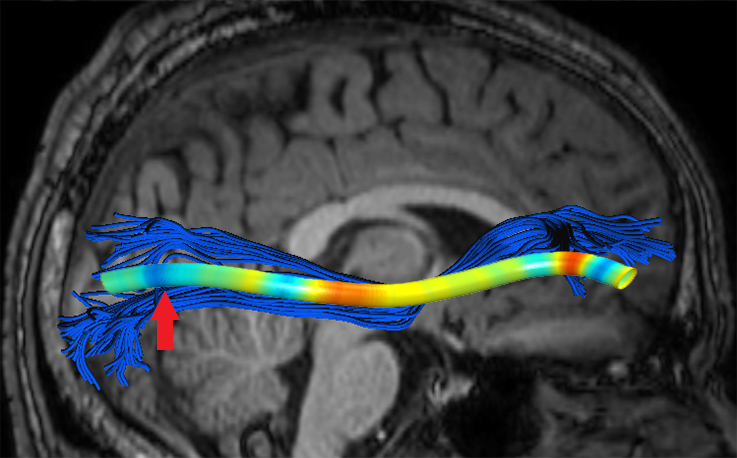}} &  & \tabularnewline
		\end{tabular}
		\par\end{centering}
	\caption{Comparison between groupwise statistical analysis based on FA-FFDD
		(top row) and standard FA (bottom row) for the left and right IFOF.
		(a) Pointwise average profile of post-season football group compared
		to normative values ($\pm1$ STD from standardized profile of NCs).
		The \textbf{\textcolor{red}{red arrow}} points to the area in which
		FA-FFDD presents improved sensitivity in comparison to standard FA.
		(b) Deviation from NCs is color-coded along the tract (in units of \#STDs from standardized
		profile).\label{FFDDvsFA}}
\end{figure}

\begin{figure}[t!]
	\begin{centering}
		\setlength{\tabcolsep}{-1pt}
		\par\end{centering}
	\begin{centering}
		\begin{tabular}{>{\centering}p{2.6cm}cc|>{\centering}p{2.6cm}cc}
			\multicolumn{2}{c}{Left IFOF} &  & \multicolumn{2}{c}{Right IFOF} & \tabularnewline
			\hline 
			{\scriptsize{}(a) Football vs. NCs} & \multicolumn{2}{c|}{{\scriptsize{}(b) Pointwise differences}} & {\scriptsize{}(a) Football vs. NCs} & \multicolumn{2}{c}{{\scriptsize{}(b) Pointwise differences}}\tabularnewline
			\includegraphics[width=2.8cm,height=2.2cm]{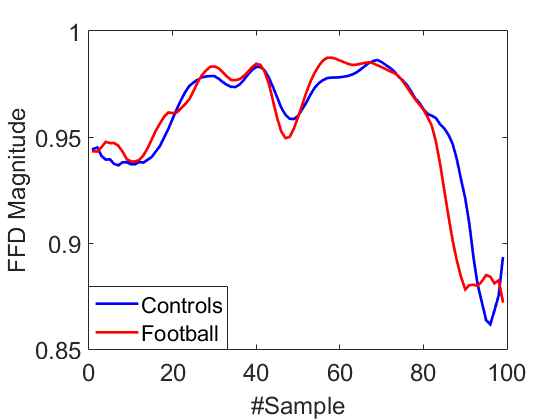} & \multirow{1}{*}[1.8cm]{\includegraphics[width=3cm,height=1.9cm]{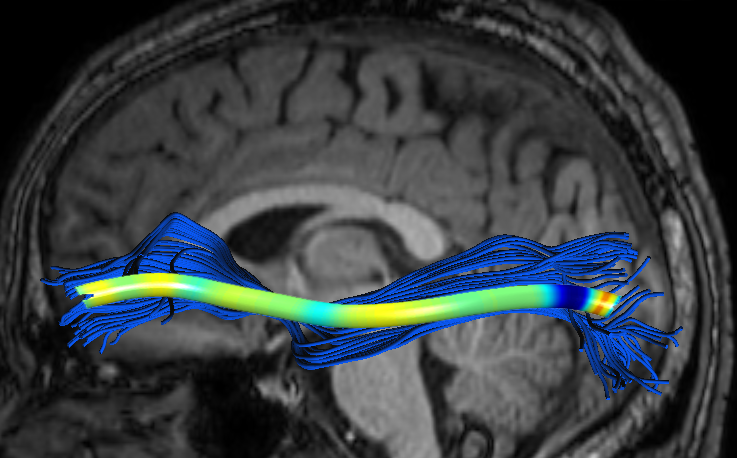}} & \multirow{1}{*}[1.8cm]{\includegraphics[height=1.9cm]{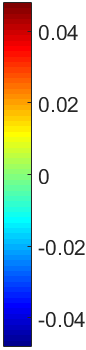}} & \includegraphics[width=2.8cm,height=2.2cm]{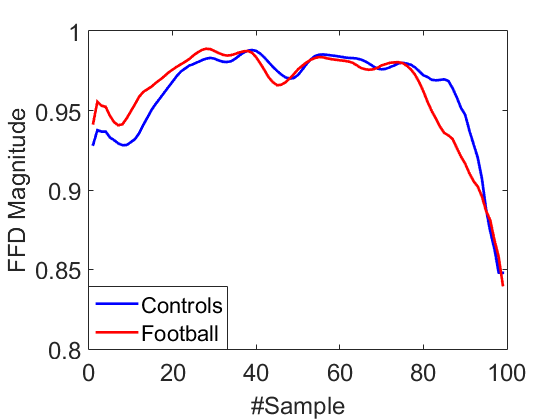} & \multirow{1}{*}[1.8cm]{\includegraphics[width=3cm,height=1.9cm]{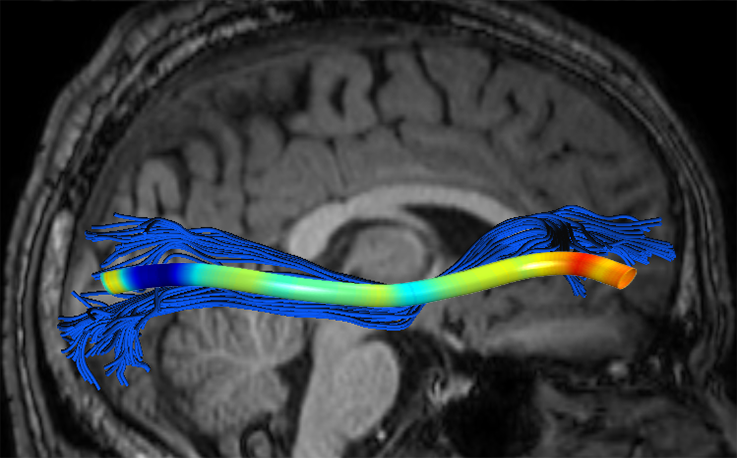}} & \multirow{1}{*}[1.8cm]{\includegraphics[height=1.9cm]{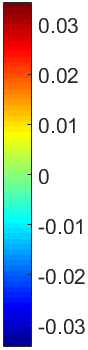}}\tabularnewline
		\end{tabular}
		\par\end{centering}
	\caption{Groupwise statistical analysis based on geometric fiber-flux density
		(FFD) alone, with no diffusion measurements, 
		for the left and right IFOF. (a) Pointwise comparison of within-group
		average FFD profiles between post-season football players and NCs.
		(b) Pointwise differences between the two profiles are color-coded along
		the tracts. The comparison shows increased flux-density variability between the
		groups at the occipital part of the tracts, allowing for increased
		sensitivity of the FFDD analysis in this area. \label{onlyFlux}}
\end{figure}

\vspace{-0.1cm}
In order to demonstrate the improved sensitivity of the proposed method in anomalies detection, we compared our FFDD groupwise analysis to an existing approach of along-tract analysis based on standard FA measurements. Similar to the FFDD analysis, normative values of standard FA were obtained by computing the pointwise mean and STD along the aligned FA profiles of NCs. The average FA profile of the football players group is then compared to these normative values. Fig. \ref{FFDDvsFA} presents a comparison between standard FA analysis and FA-FFDD analysis for the left and right IFOF.
The comparison shows that while both methods yield similar results at the frontal and central parts of the tracts, the FA-FFDD analysis shows higher variation from NCs (1 STD) compared to standard FA (0.5 STD) at the occipital part of the tracts. Note that this finding is demonstrated symmetrically for the left and right IFOF, at a spatially-consistent location with the cross-sectional and longitudinal results presented earlier in this section. 
The improved sensitivity of FA-FFDD in this region over standard FA is due to the additional geometric information provided by the proposed descriptor, as demonstrated in Fig.~\ref{onlyFlux}, which shows the group-average fiber-flux density (FFD, no coupling with diffusion measurements) of football players and NCs along the left and right IFOF. Note that higher FFD variability between the groups is indeed located at the occipital part of the tracts.

\section{Summary and Conclusion}
\vspace{-0.1cm}
We presented a novel concept of FFDD descriptors that combine geometrical
and diffusivity properties of WM fiber bundles, for local quantification
of pairwise and group-wise differences. A sub-voxel alignment of tract
profiles is accomplished by considering local FFDD dissimilarities
as an FMM inverse speed map. This allows the construction of bundle-specific
atlases for statistical analysis. Our method is demonstrated on two
datasets of contact-sports players, revealing local WM tract anomalies. 
In a group-wise comparison between active football players and normal (non-players) controls, our method revealed statistically significant differences between the groups, at spatially-consistent areas within the IFOF and FMT tracts. Furthermore, our method presented improved sensitivity to subtle structural anomalies in football players compared to along-tract FA analysis. 
The obtained results suggest the proposed method as a promising tool for mTBI assessment and localization.

\subsubsection*{Acknowledgment}
{\small This research is partially supported by the Israel Science Foundation (T.R.R. 1638/16) and the IDF Medical Corps (T.R.R.).}
\bibliographystyle{plain}
\bibliography{refMICCAI}

\end{document}